\documentclass[twoside,11pt]{article}
\usepackage{jair, rawfonts}

\usepackage{amsmath}

\usepackage{acronym}
\usepackage{amsfonts}
\usepackage{amssymb}
\usepackage{dsfont}
\usepackage{textcomp}
\usepackage[T1]{fontenc} % bold small caps
\usepackage[nameinlink]{cleveref}
\usepackage{siunitx}
\usepackage{mathrsfs}
\usepackage{vector}
\renewcommand{\vec}[1]{\vect{#1}}
\usepackage{mathtools}

\DeclarePairedDelimiterX{\infdivx}[2]{(}{)}{%
  #1\;\delimsize\|\;#2%
}

\usepackage{tikz}
\usetikzlibrary{decorations.pathmorphing} % for MCTS
\usetikzlibrary{decorations.markings} % for MCTS
\usetikzlibrary{arrows,fit} % for problem diagram
\usetikzlibrary{positioning} % for problem diagram
\usetikzlibrary{arrows.meta,calc,shapes} % for problem diagram

\usepackage{longtable,tabularx,booktabs}
\usepackage[flushleft]{threeparttable}
\usepackage{multirow}

\usepackage{pifont}% http://ctan.org/pkg/pifont
\newcommand{\xmark}{$\cdot$}%\ding{53} or \ding{5} for X-mark

\DeclareMathOperator*{\argmax}{arg\,max} % thin space
\DeclareMathOperator*{\argmin}{arg\,min} % thin space

\newcommand{\staliro}{{\scshape{S-TaLiRo}}}
\newcommand{\falstar}{{\scshape{FalStar}}}
\newcommand{\rrtrex}{{\scshape{RRT-Rex}}}

\let\osubparagraph\subparagraph
\renewcommand{\subparagraph}[1]{\osubparagraph{\indent#1}}

\usepackage{algorithm}
\usepackage{algorithmicx}
\usepackage[noend]{algpseudocode}

\usepackage[style=apa,natbib=true,doi=false,url=false]{biblatex}
\AtEveryBibitem{
\ifentrytype{inproceedings}{
 \clearlist{address}
 \clearlist{publisher}
 \clearname{editor}
 \clearlist{organization}
 \clearfield{url}  
 \clearfield{doi}  
 \clearfield{pages}  
 \clearlist{location}
 \clearlist{month}
 }{}
 
 \ifentrytype{article}{
 \clearlist{address}
 \clearlist{publisher}
 \clearname{editor}
 \clearlist{organization}
 \clearfield{url}  
 \clearfield{doi}  
 \clearlist{location}
 }{}
 }

\newacro{ai}[AI]{artificial intelligence}
\newacro{cps}[CPS]{cyber-physical system}
\newacro{stl}[STL]{signal temporal logic}
\newacro{nmac}[NMAC]{near mid-air collision}
\acrodefindefinite{nmac}{an}{a}

\jairheading{72}{2021}{377--428}{02/2021}{10/2021}

\ShortHeadings{Algorithms for Black-Box Safety Validation}{Corso, Moss, Koren, Lee, Kochenderfer}
\firstpageno{377}

\title{A Survey of Algorithms for Black-Box Safety Validation of Cyber-Physical Systems}
\author{\name Anthony Corso \email acorso@stanford.edu \\
       \addr Aeronautics and Astronautics,
       Stanford University,\\
       Stanford, CA 94305, USA
       \AND
       \name Robert J.\ Moss \email mossr@cs.stanford.edu \\
       \addr Computer Science,
       Stanford University,\\
       Stanford, CA 94305, USA
       \AND
       \name Mark Koren \email mkoren@stanford.edu \\
       \addr Aeronautics and Astronautics,
       Stanford University,\\
       Stanford, CA 94305, USA
       \AND
       \name Ritchie Lee \email ritchie.lee@nasa.gov \\
       \addr NASA Ames Research Center,\\
       Moffett Field, CA 94035, USA
       \AND
       \name Mykel J.\ Kochenderfer \email mykel@stanford.edu \\ \addr Aeronautics and Astronautics,
       Stanford University,\\
       Stanford, CA 94305, USA}

\date{March 2020}

\begin{document}

\maketitle

\begin{abstract}%
    Autonomous cyber-physical systems (CPS) can improve safety and efficiency for safety-critical applications, but require rigorous testing before deployment. 
    The complexity of these systems often precludes the use of formal verification and real-world testing can be too dangerous during development. Therefore, simulation-based techniques have been developed that treat the system under test as a black box operating in a simulated environment. Safety validation tasks include finding disturbances in the environment that cause the system to fail (falsification), finding the most-likely failure, and estimating the probability that the system fails. Motivated by the prevalence of safety-critical artificial intelligence, this work provides a survey of state-of-the-art safety validation techniques for CPS with a focus on applied algorithms and their modifications for the safety validation problem. We present and discuss algorithms in the domains of optimization, path planning, reinforcement learning, and importance sampling. Problem decomposition techniques are presented to help scale algorithms to large state spaces, which are common for CPS. A brief overview of safety-critical applications is given, including autonomous vehicles and aircraft collision avoidance systems. Finally, we present a survey of existing academic and commercially available safety validation tools.
\end{abstract}

% \begin{keywords}
%   Safety Validation, Falsification, Optimization, Path Planning, Reinforcement Learning, Cyber-Physical Systems % TODO: Keywords.
% \end{keywords}

%%%%%%%%%%%%%%%%%%%%%%%%%%%%%%%%%%%%%%%%%%%%%%%%%%%%%%%%%%%%%%%%%%%%%%%%%%%%%%%%%%%%%%%%%%%%%%%%%%
\section{Introduction}
\label{sec:intro}
Increasing levels of autonomy in cyber-physical systems (CPS) promise to revolutionize industries such as automotive transportation~\citep{DOT2018preparing} and aviation~\citep{FAA2019forecast,Kochenderfer2012next} by improving convenience and efficiency while lowering cost.  Innovations have been driven by recent progress in artificial intelligence, particularly in machine learning~\citep{LeCun2015,Russell2020} and planning~\citep{dmubook,sutton2018reinforcement}.  Machine learning has recently achieved human-competitive performance in board games~\citep{silver2017mastering,silver2016mastering}, video games \citep{Vinyals2019,mnih2015human}, and visual perception~\citep{Pillai2019,He2017}.  However, applying machine learning technologies to safety-critical domains has been challenging. Safety-critical CPS differ from conventional autonomous systems in that their failures can have serious consequences, such as loss of life and property.  As a result, these systems must undergo extensive validation and testing prior to certification and deployment. 
%TODO: Find better reference here

% Definition of safety validation challenges
Safety validation is the process of ensuring the correct and safe operation of a system operating in an environment. Desired safety properties are stipulated in a specification language and a failure is any violation of that specification. Typically, simulation is used to find failures of a system caused by disturbances in the environment, and a model of the disturbances can then be used to determine the probability of failure. A system is deemed safe if no failure has been found after adequate exploration of the space of possible disturbances, or if the probability of failure is found to be below an acceptable threshold. The procedure of proving that a system is safe to all disturbances is known as formal verification~\citep{Clarke2018,Katoen2016,Platzer2008,Fitting2012,Schumann2001} and is outside the scope of this survey.

% Description of CPS and the corresponding challenges
In this paper, we focus on CPS, which involve software and physical systems interacting over time.  This broad definition includes systems such as robots, cars, aircraft, and planetary rovers. There are several reasons why validating cyber-physical systems is challenging.  First, many of these systems contain complex components, including those produced by machine learning. The safety properties of these systems may not be well-understood and subtle and emergent failures can go undetected~\citep{yeh2018autonomous}.  Second, many systems, such as autonomous cars and aircraft, interact with complex and stochastic environments that are difficult to model.  Third, safety properties are generally defined over both the system under test and its environment. For example, the requirement that ``the test vehicle shall not collide with pedestrians'' involves both the system under test (test vehicle) and actors in its environment (pedestrians).  As a result, safety validation must be performed over the combined system. Another challenge is that sequential interactions between the system and the environment means that failure scenarios are trajectories over time, and therefore the search space is combinatorially large.  Finally, safety validation is often applied to mature safety-critical systems later in development, where failures can be extremely rare. 

% white box methods
Traditional methods for ensuring safety---through safety processes, engineering analysis, and conventional testing---though necessary, do not scale to the complexity of next-generation systems.  Advanced validation techniques are needed to build confidence in these systems. 
Many validation approaches have been proposed in the literature.  They can be broadly categorized by the information they use for analysis.  White-box methods use knowledge of the internals of the system.  For example, formal verification in the form of model checking~\citep{Clarke2018,Katoen2016} and automated theorem proving~\citep{Platzer2008,Fitting2012,Schumann2001}, represents the system using mathematical models.  Because the model is known, formal verification methods can find failure examples when they exist or prove the absence of failures when they do not.  However, because formal verification considers all execution possibilities, it often has difficulty scaling to large problems (see \cite{alur2015principles} for a discussion of verification applied to CPS).  

% Black box methods
In contrast to white-box methods, black-box techniques do not assume that the internals of the system are known. They consider a general mapping from input to output that can be sampled. Black-box methods can be applied to a much broader class of systems because they do not require a system specification.  Although model-checking techniques can be applied to some black-box systems~\citep{peled1999black}, a prohibitively large (or possibly infinite) number of samples may be required to provide complete coverage and prove the absence of failures. Instead, black-box methods often aim to quickly and efficiently find failure examples.  If no failures are found, confidence in the safety of the system will increase with additional sampling. Due to its flexibility and scalability, black-box validation is often the only feasible option for large complex systems, and is the focus of this survey.

% The safety validation tasks
We consider three safety validation tasks for a system with safety properties. First, falsification aims to find an example disturbance in the environment that causes the system to violate the property.  This formulation is useful for discovering previously unknown failure modes and finding regions where the system can operate safely. The second safety validation task is to find the most-likely failure according to a probabilistic model of the disturbances. The model can be created through expert knowledge or data to reflect the probabilities in the real environment.  The third safety validation task is to estimate the probability that a failure will occur. Failure probability estimation is important for acceptance and certification.

% Description of the types of algorithms to be used
There are many algorithms that have been used for these safety validation tasks. This survey categorizes and presents many of them. Falsification and most-likely failure analysis are related tasks in that they involve finding failures of an autonomous system. Categories of algorithms that are suited for these tasks include optimization, path planning, and reinforcement learning. Optimization approaches seek to find a trajectory of disturbances that cause the system to fail. Path planning approaches use the environment's state to aid in the exploration of possible failure modes. Reinforcement learning frames the problem as a Markov decision process and searches for a policy that maps environment states to disturbances that cause the system to fail. When the goal is to estimate the probability of failure and failures are rare, importance sampling techniques generate scenarios and translate their results into a probability estimate.  For all of the safety validation tasks, a major challenge is scalability, so problem-decomposition techniques are presented that can allow for better scalability of the presented algorithms.

% Overview of the rest of the paper. 
This paper is organized as follows. \Cref{sec:prelims} introduces notation, formally defines common terms such as safety validation or black box, and formulates three safety validation tasks. \Cref{sec:overview} gives an overview of the safety validation process, which involves defining safety properties, choosing a cost function and algorithm, and determining when sufficient testing has been performed. \Cref{sec:optimization} summarizes the optimization-based algorithms for safety validation. \Cref{sec:path_planning} describes how to use the environment state to find failures of the system through several path-planning algorithms. \Cref{sec:reinforcement_learning} shows how to apply reinforcement learning to safety validation. \Cref{sec:importance_sampling} introduces importance sampling algorithms for estimating the probability of failure. \Cref{sec:decomp} presents several ways to address problem scalability through decomposition. \Cref{sec:applications} surveys the various applications and discusses common strategies and adaptations of approaches for each domain. Finally, \cref{sec:tools} surveys existing tools in the literature and compares their basic features. 
\section{Preliminaries}
\label{sec:prelims}
This section first describes the notation used for the description of safety validation algorithms. It then defines \emph{safety validation} and \emph{black box}, and then summarizes several safety validation tasks.

\subsection{Notation}
\label{subsec:notation}

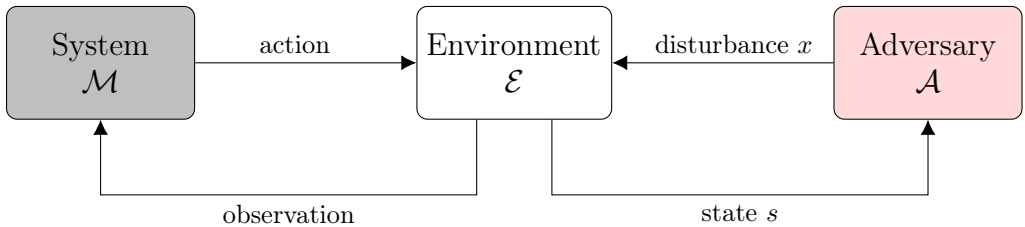
\begin{figure}[!t]
\centering
% TikZ diagram for black-box safety validation problem formulation.

\tikzset{
    >={Latex[width=2mm,length=2mm]},
        base/.style = {rectangle, rounded corners, draw=black,
        minimum width=1cm, minimum height=1cm,
        text centered},
    block/.style = {base, minimum width=2.5cm, minimum height=1.5cm},
    sutstyle/.style = {block, fill=gray!50},
    envstyle/.style = {block, fill=white}, % green!15
    advstyle/.style = {block, fill=red!15},
}

\begin{tikzpicture}
    [
        node distance=5.5cm,
        every node/.style={font=\large},
        align=center
    ]

    \node (system) [sutstyle] {System\\ $\mathcal{M}$}; % System
    \node (environment) [envstyle, right of=system] {Environment\\ $\mathcal{E}$}; % Environment
    \node (adversary) [advstyle, right of=environment] {Adversary\\ $\mathcal{A}$}; % Adversary

    \draw[->] (system) -- (environment) node [pos=0.45,above] {{\small action}};
    \draw[->] (adversary) -- (environment) node [pos=0.45,above] {{\small disturbance $x$}};
    \draw[->] (environment.south) ++(-0.5,0) -- +(0,-1) -| node[pos=0.25,below] {{\small observation}} (system);
    \draw[->] (environment.south) ++(0.5,0) -- +(0,-1) -| node[pos=0.25,below] {{\small state $s$}} (adversary);
\end{tikzpicture}
\caption{Model of the safety validation problem.}
\label{fig:problem}
\end{figure}

% System, environment, states, and safety property
A safety validation problem (\cref{fig:problem}) consists of a system $\mathcal{M}$, an environment $\mathcal{E}$, and a safety property $\psi$ that the system should have. The safety property is defined over state trajectories $\vec{s} = [s_1, \ldots, s_t]$, where $s_t \in S$ is the state of the environment at time $t \in \{1, \ldots, t_{\rm max} \}$. If the state trajectory $\vec{s}$ satisfies property $\psi$ we write $\vec{s} \in \psi$, and write $\vec{s} \not \in \psi$, otherwise. 

% Adversary, disturbances, disturbance models
The environment is perturbed by an adversary $\mathcal{A}$ through disturbances $x \in X$, where disturbances are chosen in order to induce behavior in the system which violates the safety property. Disturbance trajectories $\vec{x}$ can be chosen freely by the adversary, but they have an associated probability density $p(\vec{x})$, $p(x)$, or $p(x \mid s)$, which models their likelihood in the environment. The disturbance likelihood can be constructed through expert knowledge or learned from data.

% Simulator dynamics.
The environment transitions between states according to a dynamics function $f$ that depends on the system, environment, and disturbances. In this work, we assume that the environment and the system are fixed, making disturbances the only way to affect the system. Therefore, $f$ maps disturbance trajectories into state trajectories
\begin{equation}
\vec{s} = f(\vec{x}) \text{.}
\end{equation}
Some algorithms require the ability to simulate a disturbance $x_t$ for a single timestep from a state $s_t$, denoted
\begin{equation}
    s_{t+1} = f(s_t, x_t) \text{.}
\end{equation}

% Problem formulation diagram

\subsection{Definitions}

\paragraph{Safety Validation.}
A \emph{safety} property specifies that a certain ``bad event'' will not occur. In contrast, a \emph{liveness} property specifies that a certain ``good event'' will eventually occur. The safety-liveness distinction is important because a safety property can be shown to be violated with a concrete counterexample (the primary goal of the surveyed algorithms), while the violation of a liveness property requires formal argumentation~\citep{alpern1987recognizing}. The definition of a ``bad event'' is domain specific. 

\emph{Verification} is the process of proving that the system meets its requirements while \emph{validation} is the process of ensuring that a system fulfills its intended purpose in its operational environment~\citep{hirshorn2017nasa}. Although many of the algorithms presented in this survey can be applied to both verification and validation, we choose the term \emph{validation} to emphasize the focus on testing full-scale system prototypes in simulated operational environments. \emph{Safety validation} is therefore the processes of investigating the adherence of a system to a safety property in its operational domain.

\paragraph{Black-Box Assumption.}
A system is said to be a \emph{black box} if the system model $\mathcal{M}$ is not known or is too complex to explicitly reason about. In contrast, a \emph{white-box} system can be described analytically or specified in a formal modeling language, and a \emph{gray-box} system lies in between. Some white-box systems may be treated as a black box if knowledge of their design does not help the validation process. For example, while small neural networks can have properties formally verified by analyzing the network weights~\citep{katz2017reluplex}, large neural networks with millions or billions of parameters are generally too large for such techniques, and they would need to undergo black-box validation. In some cases, both the system and the environment are treated as a black-box, which precludes the use of validation algorithms that require the environment state (see \cref{sec:summary_of_algs} for more details).

\subsection{Safety Validation Goals}
\label{subsec:problem_formulation}
Three safety validation tasks are considered in this work and are defined below. 
\paragraph{Falsification.}
Falsification is the process of finding a disturbance trajectory that causes the outputs to violate a specification $\psi$. Such a trajectory is known as a \emph{counterexample}, \emph{failure trajectory}, or \emph{falsifying trajectory}.  Falsification finds
\begin{equation}
     \vec{x} \ \text{ s.t. } \  f(\vec{x}) \not \in \psi \text{.}
\end{equation} 

\paragraph{Most-Likely Failure Analysis.}
Most-likely failure analysis tries to find the failure trajectory with maximum likelihood
\begin{equation}
\argmax_{\vec{x}}  p(\vec{x}) \ \text{ s.t. } \ f(\vec{x}) \not \in \psi\text{.}
\end{equation}

\paragraph{Failure Probability Estimation.}\label{par:prob_est}
Failure probability estimation tries to compute the probability that a specification will be violated. Failure probability is given by the expectation of observing a failure under the disturbance model
\begin{equation}
    P_{\rm fail} = \mathbb{E}\bigl[ \mathds{1}{\{ f(\vec{x}) \not \in \psi \}} \bigr]\text{.}
\end{equation}

\section{Overview of Solution Techniques}
\label{sec:overview}
Solving a safety validation problem for a given CPS requires the following steps: 1) define a safety property to validate, 2) define an appropriate cost function to guide the search, 3) choose a safety validation algorithm, which depends on the system, environment and safety validation task and 4) run it until a counterexample is discovered (for falsification and most likely failure analysis) or the space of possible scenarios has been sufficiently covered. This section provides an overview of each of these steps.

\subsection{Safety Specification with Temporal logic}
In safety validation, the first step is to define a safety property to validate. Although safety properties could be defined using natural language or human preferences (both of which are highly expressive), formal specification languages are often preferred because they reduce ambiguity and permit efficient numerical evaluation of state trajectories. The most common formal specification languages are based on temporal logic.

Temporal logic is a logical framework for describing properties of signals over time.  It enables reasoning about time and temporal information.
Properties are stated as formulas that evaluate to a Boolean value and the syntax of these formulas is governed by a grammar. 
Various temporal logics have been proposed for different domains, including linear temporal logic~\citep{pnueli1977temporal}, metric temporal logic~\citep{koymans1990specifying}, and computation tree logic~\citep{clarke1981design}.
Signal temporal logic (STL)\acused{stl} is a temporal logic for real-valued signals that is widely used as a specification language for the safety validation of cyber-physical systems~\citep{Donze2010robust,kapinski2016simulation}.  The basic unit of an \ac{stl} formula is an atomic formula of the form $\mu(x_i) \geq 0$, where $x_i$ is a real-valued signal and $\mu$ is an arbitrary function from $\mathbb{R}^{n}$ to $\mathbb{R}$.
A combination of Boolean and temporal operators can be applied to one or more atomic formulas to form more complex formulas.  Boolean operators can include unary or binary Boolean operators, such as \emph{negation} $\neg$, \emph{conjunction} $\wedge$, and \emph{disjunction} $\vee$.  Temporal operators reason over the temporal aspect of signals. Examples of temporal operators include \emph{always} $\square \psi$ ($\psi$ holds on the entire subsequent path), \emph{eventually} $\lozenge \psi$ ($\psi$ holds somewhere on the subsequent path), and \emph{until} $\psi_1 \mathcal{U} \psi_2$ ($\psi_1$ holds at least until $\psi_2$ becomes true on the subsequent path).  Temporal operators may be indexed by a time interval $I$ 
%of $\mathbb{R}^{+}$ 
over which the property is considered.  

The following are some examples of \ac{stl} formulas representing safety properties: 
\begin{align}
\psi_1: \qquad & \lozenge_{[0,100]}(d_{\text{goal}} < 10)\label{eq:stl_spec1}\\
\psi_2: \qquad & \square_{[0,\infty]}(\neg(d_{h} < 500 \wedge d_{v} < 100))\label{eq:stl_spec2}
\end{align}
The specification $\psi_1$ (\cref{eq:stl_spec1}) states that at some point in the next 100 seconds, the distance to the goal $d_{\text{goal}}$ becomes less than 10 feet.  Such a specification could be used, for example, to require that an autonomous vehicle reach a goal location within a certain time limit.
The specification $\psi_2$ (\cref{eq:stl_spec2}) states that for all time, the horizontal separation $d_h$ and the vertical separation $d_v$ between two aircraft shall not be simultaneously less than 500 feet and 100 feet, respectively.  This specification describes the absence of a near mid-air collision, which is an important safety event for aircraft collision avoidance systems~\citep{Kochenderfer2012next}.
% Some temporal logics such as \ac{stl} support quantitative semantics, where a formula can map a signal trace to a real value, called \emph{robustness}, which represents the degree of satisfaction of the formula.  Positive robustness values indicate that the formula is satisfied while negative values mean that the formula is not satisfied.  %Many safety validation algorithms use temporal logic robustness as a basis for the cost function.

\subsection{Cost Functions}

A naive approach to safety validation is to search randomly over disturbance trajectories until a counterexample is discovered. If counterexamples are rare, this process can be inefficient or even intractable. A cost function $c_{\text{state}}(\vec{s})$ that measures the level of safety of the system over the state trajectory $\vec{s}$ can be used to guide the search. A well-designed cost function will help bias the search over disturbance trajectories toward those that are less safe. Additionally, if the goal is to find the most-likely failure, then the cost function can incorporate the likelihood of the disturbance trajectory. 

Once a cost function is defined, safety validation becomes an optimization problem over disturbance trajectories
\begin{equation}
    \vec{x}^* = \argmin_{\vec{x}} c(\vec{x}) \text{,} \label{eq:optimization_problem}
\end{equation}
where $c(\vec{x}) = c_{\text{state}}(f(\vec{x}))$. The cost function is designed such that $c(\vec{x}) \geq \epsilon \iff f(\vec{x}) \in \psi$, where $\epsilon$ is a safety threshold. Therefore, if a disturbance $\vec{x}$ causes $c(\vec{x}) < \epsilon$, then $\vec{x}$ is a counterexample. 

The design of the cost function $c$ is specific to the application, but can be done ad-hoc or using a formal measure of the satisfaction of a safety property. For simple safety proprieties such as collision avoidance, the miss distance (point of closest approach between agents) is a common choice~\citep{koren2018adaptive,lee2020adaptive}. \citet{balkan2017underminer} propose several possible functions for finding non-convergence behaviors in control systems, including Lyapunov-like functions, $M$-step Lyapunov-like functions, neural networks, and support vector machines.

When the specification is represented by a temporal logic expression, a natural choice for $c$ is the temporal logic robustness $\rho(\vec{s}, \psi)$. The robustness is a measure of the degree to which the trajectory $\vec{s}$ satisfies the property $\psi$. Large values of robustness mean that at no point does the trajectory come close to violating the specification, while low but positive values of robustness mean that the trajectory is close to violating the specification. A robustness value less than zero means that the specification has been violated and gives an indication of by how much. The robustness for space-time signals can be computed from a recursive definition~\citep{fainekos2009robustness, Donze2010robust,yang2013dynamic,LeungArechigaEtAl2020}. The derivative of the robustness with respect to the state trajectory can be approximated, which may help derive gradient-driven coverage metrics~\citep{pant2017smooth, LeungArechigaEtAl2020} or with doing gradient-based optimization if the black-box assumption is relaxed. Causal information in the form of a Bayesian network can be incorporated into the robustness for improved falsification~\citep{akazaki2017causality}. 
Connections have also been made between robustness and delta-reachability to relate falsification and exhaustive search through approximations~\citep{abbas2017relaxed}.

Robustness can be a challenging objective to optimize because it can be non-smooth~\citep{LeungArechigaEtAl2020} and be dominated by state variables with larger magnitude values. \citet{LeungArechigaEtAl2020} propose smooth approximations to the robustness and \citet{Akazaki2018falsification} optimize a convex function of the robustness $-\textrm{exp}\left(-\rho(\vec{s})\right)$, which bounds the maximum cost. State variables may be normalized to alleviate differences in scale between the variables, but normalization requires the range of the state variables which may not be known a priori. \citet{zhang2019multi} propose measuring the robustness of each state variable independently and using a multi-armed bandit algorithm to decide which robustness value to optimize on each iteration.

% TODO flip the order of presentation and explain more
In addition to measuring safety, cost functions may include other search heuristics. For most likely failure analysis, the cost function includes the likelihood of the disturbance trajectory $p(\vec{x})$ for counterexamples
\begin{equation}
    c^\prime(\vec{x}) = \begin{cases}
        c(\vec{x})  & {\rm if} \ c(\vec{x}) \geq \epsilon \\
        -p(\vec{x}) & {\rm if} \ c(\vec{x}) < \epsilon \text{.}
    \end{cases}
\end{equation}
An objective that achieves a similar goal but may be easier to optimize includes a penalty term for low-likelihood disturbances
\begin{equation}
    c^\prime(\vec{x}) = c(\vec{x}) - \lambda p(\vec{x}) \text{,}
\end{equation}
where the penalty $\lambda > 0$ is user-defined. Additional penalty terms related to coverage (see \cref{sec:coverage} for an overview) can be included to encourage exploration. Domain-specific heuristics are also possible. For example, \citet{qin2019automatic} penalize disturbances that do not follow a domain-specific set of constraints, which is useful for getting adversarial agents in a driving scenario to follow traffic laws. Other examples of domain-specific heuristics are described in \cref{sec:applications}.

Solution techniques that build trajectories sequentially (such as those based on planning and reinforcement learning) can require the evaluation of trajectory segments or states. Upper and lower bounds on robustness can be computed for incomplete trajectory segments~\citep{dreossi2015efficient}. An approach for most-likely failure analysis called adaptive stress testing (AST)~\citep{lee2020adaptive,koren2019adaptive} defines a cost for individual state-disturbance pairs
\begin{equation}
    c(s_t, x_{t}) = \begin{cases}
        0  & {\rm if} \ s_t \in S_{\rm fail} \\
        \lambda & {\rm if} \ s_t \not \in S_{\rm fail}, \ t \geq t_{\rm max} \\
        \log p(x_t \mid s_t) & {\rm if} \ s_t \not \in S_{\rm fail}, \ t < t_{\rm max} \text{,}
    \end{cases}
\end{equation}
where the specification $\psi$ is for the system to avoid reaching a set of failure states $S_{\rm fail}$, and $\lambda$ penalizes disturbances that do not end in a failure state. The log probability of the disturbance is awarded at each time step to encourage the discovery of likely failures.

\subsection{Overview of Algorithms}
\label{sec:summary_of_algs}

As demonstrated in the previous section, safety validation algorithms solve an optimization problem to discover counterexamples. This section outlines four categories of algorithms distinguished by the information they use for optimization, the requirements on the simulated environment, and the desired safety validation task. For each category, we summarize the approach and describe its strengths, weaknesses, and requirements. Specific algorithms are discussed in more detail in \cref{sec:optimization,sec:path_planning,sec:reinforcement_learning,sec:importance_sampling,sec:decomp}.

% Each category of safety validation algorithm has benefits and drawbacks that depend on safety validation task and the details of the system and environment. We allude to some of these strengths and weaknesses in the preceding sections but here we try to spell them out with clarity

% %TODO: Integrate
% From background: Adversaries that use the system state require an environment that is constructed to provide that information. In cases where the state information is unavailable (possibly due to the simulator implementation or privacy concerns), the adversary must produce disturbance trajectories based solely on the outcomes of past trajectories. 

\subsubsection{Black-Box Optimization.} 

Many safety validation algorithms attempt to solve \cref{eq:optimization_problem} directly over the space of disturbance trajectories. Although gradient-based approaches cannot be used due to the black-box assumption on the system, there are many existing algorithms that can be applied with no modification~\citep{kochenderfer2019algorithms}.  Due to the complexity of many autonomous systems and environments, however, the optimization problem is generally non-convex and can have many local minima. Therefore, algorithms that can escape local minima and have adequate exploration over the space of disturbance trajectories are preferred. 

Two algorithms that have been used out of the box in falsification software~\citep{annapureddy2011staliro,donze2010breach} (see \cref{sec:tools} for more details) are covariance matrix adaptation evolution strategy (CMA-ES)~\citep{hansen1996adapting} and globalized Nelder-Mead (GNM)~\citep{luersen2004globalized}, both of which are effective for global optimization. Another way to escape local minima is to combine global and local search~\citep{deshmukh2015stochastic,adimoolam2017classification, yaghoubi2019gray,Mathesen2019falsification}, where one optimization routine is used to explore the space of disturbances trajectories to find regions of interest, and another algorithm does local optimization to find the best disturbance trajectory in a region.

The primary benefit of optimization-based safety validation is the minimal set of restrictions placed on the simulator of the system and the environment. Black-box optimizers do not need access to the state of the environment and only needs to return the value of a safety metric for a given disturbance trajectory. The simulation state may be unavailable for implementation or privacy reasons so optimization-based approaches would be a good choice in those cases. If, however, the state is available and would be useful for finding failures, then optimization approaches may not function as well as path planning or reinforcement learning approaches. If an environment has stochasticity in state transitions (in addition to the applied disturbances), then optimization techniques such as Bayesian optimization can be used to account for it. The biggest drawback to optimization strategies is the need to optimize over the entire space of disturbance trajectories, which scales exponentially with the time horizon.

\subsubsection{Path Planning.} 

Safety validation may be framed as a planning problem through the state space of the environment using the disturbances as control inputs. Planning algorithms construct a disturbance trajectory $\vec{x}$ from an initial state $s_0$ to a set of failure states $S_{\rm fail}$, with a corresponding cost for each transition. Classical planning algorithms~\citep{ghallab2004automated} typically make the assumption that there exists a model in a formal language such as PDDL~\citep{mcdermott1998pddl} or STRIPS~\citep{fikes1971strips}, and that the size of the state space is discrete and not too large. Both assumptions are violated when dealing with black-box autonomous systems operating in continuous state spaces so traditional planning algorithms such as forward and backward chaining cannot be directly applied.  Other planning algorithms such as value iteration~\citep{sutton2018reinforcement}, Dijkstra's algorithm~\citep{dijkstra1959note}, and their variants can work with a black-box transition function but do not scale well to large or continuous state spaces. Heuristics such as state novelty~\citep{lipovetzky2012width} have been successfully applied to black-box planning settings with large state spaces~\citep{lipovetzky2015classical,frances2017purely}, and could therefore be applicable to safety validation of CPS. 

Planning problems with continuous and high-dimensional state spaces commonly occur in the field of robotics and path planning (see the overview by \citet{lavalle2006planning}). Optimal control algorithms~\citep{lewis2012optimal} are common in this domain but typically violate the black-box assumption by requiring gradients of the transition function. Sampling based approaches such as the rapidly exploring random tree (RRT) algorithm and the probabilistic roadmap planner~\citep{kavraki1996probabilistic} can work with a black-box simulator and have been shown to be effective in high-dimensional continuous planning problems. The probabilistic roadmap planner, however, requires the ability to connect independent state trajectory segments making it difficult to use in a completely black-box setting. RRT on the other hand, grows a search tree forward from the set of reachable states and can therefore be used with a black-box simulator. RRT has been used extensively for safety validation of CPS and variations on the algorithm will be discussed in greater detail in \cref{sec:path_planning}.

Path planning algorithms rely heavily on the environment state to discover failures, which can be a strength and weakness. Planning algorithms can efficiently search high-dimensional state spaces by reusing trajectory segments, and often provide a natural way to compute state space coverage which can be used to determine when sufficient testing has been done. If the reachable set of states is small compared to the state space, however, planning algorithms may perform poorly~\citep{kim2005rrt}. A drawback to path planning algorithms is their inability to naturally handle stochasticity. Most path planning algorithms rely on the ability to deterministically replay trajectory segments or initialize a simulator into a predefined state, which may be challenging for some simulators.  Additionally, if the problem has a long time horizon, then prohibitively large trees may be required to find failure trajectories.

\subsubsection{Reinforcement Learning.}
In reinforcement learning (RL), the safety validation is modeled as a Markov decision process (MDP). An MDP is defined by a transition model $P(s^\prime \mid s, x)$ that gives the probability of arriving in state $s^\prime$ given the current state $s$, a disturbance (or action) $x$, a reward $R(s, x)$, and a discount factor $\gamma \in [0,1]$ that decreases the value of future rewards. RL algorithms learn  a policy $\pi$ (a function that maps states to disturbances $x = \pi(s)$) that maximizes expected future rewards. Typically the reward function is chosen to be $R(s,x) = - c(s,x)$ so that maximizing reward minimizes cost. For an overview of MDPs and their solvers see the texts by \citet{dmubook} or \citet{sutton2018reinforcement}. 

% There are several concepts from the MDP literature that are useful for understanding the algorithms in this section. The first is the notion of the value function $V^\pi(s)$, which is the expected discounted sum of future rewards when in the state $s$ and following the policy $\pi$. The value function can be computed as the solution to the Bellman equation
% \begin{equation}
%     V^\pi(s) = \mathbb{E}\left[ R\left(s, \pi(s)\right) + \gamma V^\pi(s^\prime) \right] \text{.}
% \end{equation}
% The optimal policy $\pi^*$ is defined as 
% \begin{equation}
%     \pi^*(s) = \argmax_{\pi} V^\pi(s) \text{.}
% \end{equation}
% The state-action value function for a policy is $Q^\pi(s, x)$ which is the value of being in state $s$, applying disturbance $x$, and then following the policy $\pi$. It is defined by the Bellman equation
% \begin{equation}
%     Q^\pi(s,x) = \mathbb{E}\left[ R(s, x) + \gamma Q^\pi\left(s^\prime, \pi(s^\prime)\right) \right]\text{.}
% \end{equation}
% If the size of the state space and disturbance space is discrete then $Q^\pi$ can be computed exactly with matrix inversion or dynamic programming~\citep{dmubook}. If the state or action space is large or continuous (as is often the case for cyber-physical systems), $Q^\pi$ can be estimated through random sampling and bootstrapping~\citep{sutton2018reinforcement}. 

Reinforcement learning algorithms are similar to path-planning approaches because they also rely on the environment state (unless specifically formulated otherwise as in \citet{koren2019adaptive}). While planning algorithms search for full trajectories, RL algorithms learn a policy that generates disturbances from the current state. Since policies do not explicitly represent the entire disturbance trajectory, they may be easier to optimize and can be applied to long time horizon problems. Uncontrolled stochasticity in the environment is naturally handled by RL algorithms, which are designed to function in stochastic environments. Additionally, many RL algorithms can be used with episodic simulators that only require the ability to reset and step forward in time, rather than the ability to initialize to a particular state. The downside to RL algorithms is that they may be sample inefficient and require complex (and sometimes brittle) training procedures compared to optimization and path-planning approaches.

\subsubsection{Importance Sampling.}
For many cyber-physical applications, it is impossible to design an autonomous system that never violates a safety property. In that case, failure probability is a useful metric of safety. If failure events are rare, then Monte Carlo approaches will require a large number of samples before converging to the true probability of failure \citep{hahn1972sample}. To address this problem, importance sampling approaches artificially increase the likelihood of failure with a proposal distribution $q(\vec{x})$, and then weight observed failures to get an unbiased estimate of the probability of failure with fewer samples.

In addition to causing more failures, the proposal distribution has the property that $q(\vec{x}) > 0$ everywhere that $p(\vec{x})\mathds{1}\{ c(\vec{x}) < \epsilon \} > 0$ (so all disturbances that lead to failure can be sampled from $q$). The proposal distribution is also referred to as the biased, sampled, or importance distribution. The importance sampling estimate of the probability of failure is done by taking $N$ samples drawn from $q$ and computing the weighted average
\begin{equation}
    \hat{P}_{\rm fail} = \frac{1}{N} \sum_{i=1}^N \frac{p(\vec{x}_i)}{q(\vec{x}_i)} \mathds{1}{\{ c(\vec{x}_i) < \epsilon \}} \text{.} \label{eq:is_est}
\end{equation}
The variance of the importance sampling estimate is given by 
\begin{equation}
    {\rm Var}(\hat{P}_{\rm fail}  ) = \frac{1}{N} \mathbb{E}_q \left[ \frac{(p(\vec{x})\mathds{1}{\{ c(\vec{x}) < \epsilon \}}  - q(\vec{x})P_{\rm fail})^2}{q(\vec{x})} \right] \text{.} \label{eq:is_var}
\end{equation}
The goal of a good importance sampling distribution is to minimize the variance of the estimator $\hat{P}_{\rm fail}$ so that fewer samples are needed for a good estimate. It is clear from \cref{eq:is_var} that a zero variance estimate can be obtained with the optimal importance sampling distribution
\begin{equation}
    q^*(\vec{x}) = \frac{p(\vec{x}) \mathds{1}{\{ c(\vec{x}) < \epsilon \}}} {P_{\rm fail}} \text{.}
\end{equation}
Generating this distribution is not possible in practice because $c(\vec{x})$ is a black box and the normalization constant $P_{\rm fail}$ is the very quantity being estimated. Importance sampling algorithms seek to estimate the optimal importance sampling distribution $q^*(\vec{x})$.

Importance sampling approaches require finding many failure examples to learn a distribution over failures. Therefore, failure examples can, in principle, be found using any of the three previous approaches. The most common importance sampling approaches such as multilevel splitting and the cross-entropy method function most similarly to optimization-based techniques because they search directly over the space of disturbance trajectories and do not require state information. These techniques therefore carry the same strengths and weakness as optimization techniques. Likewise, importance sampling techniques that involve the use of environment state have the same benefits and limitation as the corresponding path planning or RL algorithm. 

\subsection{Coverage Metrics}
\label{sec:coverage}

A natural question that arises in safety validation is ``when is testing complete?'' There are an infinite number of possible tests, so measures of \emph{coverage} are used as a principled way of determining when sufficient testing has been performed to declare a system safe. We overview three types of testing coverage: the probability of failure, coverage of the space of disturbance trajectories, and coverage of the reachable set of states. 

\subsubsection{Probabilistic Coverage.}
Testing of a safety critical system may be complete when the estimated probability of failure is below a specified threshold with high confidence. The outcome of each simulation of a disturbance trajectory is a Bernoulli random variable, 
\begin{equation}
    \mathds{1}\left\{ f(\vec{x}) \not \in \psi \right\}
\end{equation}
where a positive outcome (failure) occurs with probability $P_{\rm fail}$. These samples can be used with the statistical tools of hypothesis testing and estimation to determine when sufficient testing has been performed.

In hypothesis testing, two complimentary hypotheses are compared based on the evidence of sample trajectories. For example, the null hypothesis $H_0$ may be that the probability of failure is less than an acceptable threshold $p_0$,  $H_0 \coloneqq P_{\rm fail} < p_0$, and the alternative hypothesis is the complement $H_1 \coloneqq P_{\rm fail} \geq p_0$. The level of confidence in a hypothesis test is specified by the probability of type I and type II errors (with lower probability of error requiring more data). For an overview of hypothesis testing techniques for statistical model checking see the work of \citet{agha2018survey}.

% Estimation testing
Estimation techniques compute a confidence interval over the probability of failure. The frequentist approach involves taking the mean of $N$ independent samples to produce an estimate of the probability of failure
\begin{equation}
    \hat{P}_{\rm fail} = \frac{1}{N} \sum_{i=1}^N \mathds{1}\left\{ f(\vec{x}_i) \not \in \psi \right\} \text{,}
\end{equation}
and uses a concentration inequality to estimate the bounds. Suppose after $N$ samples no failures have been observed, then Hoeffding's inequality states that
\begin{equation}
    {\rm Pr}\left[ P_{\rm fail} \geq p_0 \right] \leq 2e^{-2N p_0^2} \text{.}
\end{equation}
The number of samples $N$ can be chosen so the right-hand side of the inequality is sufficiently small. In contrast, the Bayesian approach involves maintaining a belief over the probability of failure that gets updated with each new sample. For Bernoulli random variables, the Beta distribution can be used to represent the belief. Benefits of the Bayesian approach include the ability to incorporate a prior estimate of the probability of failure and the ability to explicitly compute confidence bounds. 

When validating safety-critical systems, failures are often rare and safety thresholds are strict, so many samples are required to make probabilistic arguments for safety. To alleviate this burden, importance sampling approaches are used (see \cref{sec:importance_sampling} for details), where samples are drawn from a proposal distribution $q(\vec{x})$, under which failures are more likely. The samples are then appropriately weighted when performing hypothesis tests~\citep{harrison2012conservative} or estimation. 

A poorly chosen, or \emph{inefficient}, proposal distribution can increase the variance of an estimator, which in turn reduces the confidence in the estimate. An inefficient proposal distribution can be identified when the importance weights are large, or equivalently, when the effective sample size is small compared to the actual number of samples. When samples with large weights are being drawn, \citet{kim2016improving} suggest limiting the maximum weight by clipping the proposal distribution in regions with large weights. \citet{uesato2019rigorous} suggest combining the importance sampling estimator with a basic Monte Carlo estimator to minimize the downside of a bad proposal distribution and \citet{neufeld2014adaptive} provide a principled way of choosing the best estimator from several possibilities. 

\subsubsection{Disturbance-Space Coverage.}
If failure events have extremely low probability, or if no probability model is available, coverage can be measured by how well the sampled disturbance trajectories fill the space of possible trajectories. Let $X \times T$ be the space of disturbance trajectories, and let $V$ be a set of sampled trajectories. Informally, the coverage $C(V) \in [0,1]$ is the degree to which $V$ represent the space $X \times T$. Testing can be concluded when the coverage reaches close to $1$. Although a variety of coverage metrics can be employed, \emph{dispersion}~\citep{esposito2004adaptive} and \emph{star discrepancy}~\citep{Nahhal2007Test,dang2008sensitive,dreossi2015efficient} have previously been used for safety validation.

% Dispersion
For a distance metric between states $d$, the dispersion is the radius of the largest ball in $X \times T$ that contains no points in $V$, i.e. $\max_{\vec{x} \in X \times T} \min_{\vec{x}_i \in V} d(\vec{x}, \vec{x}_i)$. Dispersion is challenging to compute in many dimensions and can be overly conservative~\citep{esposito2004adaptive}. Instead, a coverage metric based on average dispersion can be computed on a grid with $n$ points and a spacing of $\delta$ as
\begin{equation}
    C_{\rm disp}(V) = 1 - \frac{1}{\delta} \sum_{j=1}^{n} \frac{\min\left(d_j(V), \delta\right)}{n} \text{,}
\end{equation}
where $d_j(V)$ is the shortest distance from grid point $j$ to any node in $V$ according to the metric $d$~\citep{esposito2004adaptive}. The value $\min(d_j(V), \delta)$ is therefore the radius of the largest ball that can sit at grid point $j$ and not touch any nodes in the tree or another grid point. Note that $C_{\rm disp} \to 1$ when $V$ contains points at each grid point. The coverage is therefore related to the fidelity of the grid, with a finer grid giving a more comprehensive value at greater computational expense.

% Star discrepancy
The star discrepancy coverage metric measures how evenly a set of points is distributed. The discrepancy $D$ of a set over a region $B \subseteq X \times T$ is
\begin{equation}
    D(V, B) = \frac{|V \cap B|}{|V|} - \frac{{\rm vol}\left(B \right)}{{\rm vol}\left(X\times T \right)} \text{,}
\end{equation}
where $\rm vol$ is the volume of a region and $|V|$ is the number of sample points. The first term is the fraction of the points that are in $B$ and the second term is the ratio of the volume of $B$ to the volume of space of disturbance trajectories. The star discrepancy $D^*$ is the largest discrepancy over all possible subregions of $S$
\begin{equation}
    D^*(V) = \max_{B} D(V, B) \text{.}
\end{equation}
Note that $D^* \to 0$ when all subregions of $S$ have their ``fair share'' of the sample points based on volume, and $D^* \to 1$ when all the sample points overlap. To turn discrepancy into a coverage metric, define
\begin{equation}
    C_{\rm disc}(V, B) = 1 - D(V, B)
\end{equation} 
and
\begin{equation}
    C_{\rm star}(V) = 1 - D^*(V) \text{.}
\end{equation}
It is possible to approximate $D^*(V)$ using a finite partitioning of the state space into regions $\{ B_1, \ldots, B_n \}$ such that $\cup_{i=1}^n B_i  = X \times T$. The discrepancy coverage can be computed for each box $C_{\rm disc}(V \cap B_i, B_i)$~\citep{dreossi2015efficient} or a lower and upper bound of the star discrepancy coverage can be computed based on $B_i$~\citep{thiemard2001algorithm,Nahhal2007Test}.

\subsubsection{State-Space Coverage.}
When the space of disturbance trajectories is prohibitively large (e.g. for long time horizon problems), it can be more efficient to define coverage metrics for the state space. In fact, both dispersion and star-discrepancy were first used as state-space coverage metrics for the rapidly exploring random tree algorithm~\citep{esposito2004adaptive, Nahhal2007Test,dang2008sensitive,dreossi2015efficient}. 

The challenge to using state-space coverage metrics is that not every state in the state space may be reachable. A state $s$ is \emph{reachable} if there is a disturbance trajectory that can be applied that will cause the environment to reach $s$. In many safety validation problems, the reachable states are a subset of the full state space due to the limited control via disturbances~\citep{esposito2004adaptive}. If the reachable set is small compared to $S$, then the maximum possible coverage value will also be small, and testing will never terminate. To mitigate this problem, \citet{esposito2004adaptive} proposed a growth metric on the set of samples defined as
\begin{equation}
    g(V) = \Delta C(V) / \Delta |V| \text{.}
\end{equation}
The growth $g$ measures how much the coverage metric increases with the increase in sample points in $V$. In addition to stopping based on adequate coverage, testing can be terminated if the growth is below a specified threshold, suggesting that adding more sample points to $V$ will not improve the coverage. 

\section{Black-Box Optimization} 
\label{sec:optimization}

This section surveys algorithms that use black-box optimization techniques to solve safety validation problems, as well as the modifications that the techniques require to be effective for safety validation. Safety validation has been performed with simulated annealing, evolutionary algorithms, Bayesian optimization, and extended ant colony optimization. 
     
%%%%%%%%%%%%%%%%%%%%%%%%%%%%%%%%%%%%%%%%%%%%%%%%%%%%%%%%%%%%%%%%%%%%%%%%%%%%%%%%%%%%%%%%%%%%%%%%%%%%%%%%%%%%
\subsection{Simulated Annealing}
\label{sec:simulated_annealing}

An approach to stochastic global optimization known as simulated annealing uses a random walk around the disturbance space to minimize a cost function $c$. A temperature parameter $\beta$ is used to control the amount of stochasticity in the method over time and a transition function $P(\vec{x}^\prime \mid \vec{x})$ describes the probability distribution over the next disturbance trajectory $\vec{x}^\prime$. If the new trajectory is a counterexample, then it is returned, otherwise it is adopted as $\vec{x}$ with probability ${\rm exp}(-\beta(c(\vec{x}^\prime) - c(\vec{x})))$. Due to the stochastic nature of simulated annealing, it is often a suitable algorithm for solving a global optimization problem with many local minima and can therefore be effective for safety validation~\citep{abbas2013probabilistic,aerts2018temporal}.

A common choice for transition function is to use a Gaussian distribution around the current point $\vec{x}$ with a standard deviation that is adjusted based on the ratio of accepted points~\citep{kochenderfer2019algorithms}. This approach may not work well when the disturbance space has constraints that must be satisfied, such as lower and upper bounds on the possible disturbances~\citep{abbas2013probabilistic}. \citet{abbas2013probabilistic} proposes the use of a \emph{hit and run} approach to transitioning that respects constraints. It follows three steps:
\begin{enumerate}
    \item Sample a random direction $\vec{d}$ in the disturbance trajectory space.
    \item Perform a line search in the direction of $\vec{d}$ to determine the range of $\alpha$ such that $\vec{x} + \alpha \vec{d}$ does not violate any constraints.
    \item Sample $\alpha$ from this range according to a chosen distribution. The standard deviation of this distribution can be adjusted using the acceptance ratio to improve convergence. 
\end{enumerate}
\citet{aerts2018temporal} improved the hit and run scheme by suggesting that $\alpha$ be chosen for each disturbance dimension separately so that highly constrained dimensions do not restrict the step size of less constrained dimensions~\citep{aerts2018temporal}.

Typically, the size of $\vec{x}$ remains fixed throughout the optimization, meaning the temporal discretization of the disturbance trajectory is never adjusted. \citet{aerts2018temporal} note that the frequency content of the disturbance trajectory is salient for some falsification problems, and therefore the temporal discretization of the disturbance trajectory should itself be optimized. An approach called input-signal-space optimization uses a two-layered approach, where an outer loop uses SA to select the length of the disturbance trajectory, and an inner loop finds the lowest cost trajectory for that length. This approach is able to adapt the fidelity of the time domain, getting improved results for some falsification problems~\citep{aerts2018temporal}.

%%%%%%%%%%%%%%%%%%%%%%%%%%%%%%%%%%%%%%%%%%%%%%%%%%%%%%%%%%%%%%%%%%%%%%%%%%%%%%%%%%%%%%%%%%%%%%%%%%%%%%%%%%%%
\subsection{Evolutionary Algorithms} \label{sec:genetic_algorithms}

Evolutionary algorithms approach global optimization by mimicking the biological process of evolution. A population of individuals is sampled and then evolved using the processes of \emph{crossover}, where low cost individuals are combined to form new individuals, and \emph{mutation}, where individuals are randomly modified to incorporate new traits into the population.

\citet{zhao2003generating} applied a genetic algorithm to generate test cases for embedded control systems. Each individual is represented by a real-valued vector for all continuous variables and a binary encoding for all discrete variables. Selection occurs with a probability inversely related to the cost function. The crossover between two individuals is done using either a randomized concatenation of subsequences or an arithmetic combination of the individuals. Although evolutionary algorithms lack convergence guarantees and are largely based on heuristics, they can be effective for problems with very large input spaces, as demonstrated by their successful use in the safety validation of multi-agent sense and avoid algorithms~\citep{zou2014safety}. 

Evolutionary algorithms can optimize more complex inputs such as temporal logic expressions represented as trees. \citet{corso2020interpretable} hypothesize that counterexamples can often be described by a simple temporal logic property of the disturbance trajectory, called a \emph{failure description}. Failure descriptions lend interpretability to automated testing and have been shown to be effective at finding counterexamples as well as giving engineering insight into the failure modes of the system.  Genetic programming is used to optimize failure descriptions $\varphi$ that satisfy
\begin{equation}
    \vec{x} \in \varphi \implies c(\vec{x}) < \epsilon \text{.}
\end{equation}
The cost function for a given failure description is the average cost of trajectories that satisfy it. Sampling satisfying trajectories is, in general, a hard problem, but \citet{corso2020interpretable} provide an algorithm to do so under a set of assumptions.

%%%%%%%%%%%%%%%%%%%%%%%%%%%%%%%%%%%%%%%%%%%%%%%%%%%%%%%%%%%%%%%%%%%%%%%%%%%%%%%%%%%%%%%%%%%%%%%%%%%%%%%%%%%%
\subsection{Bayesian Optimization} \label{subsec:bayes_opt}
In Bayesian optimization~\citep{mockus2012bayesian}, a  surrogate model (such as a Gaussian process) is used to represent the cost function over the space of disturbance trajectories. The model is used to select disturbance trajectories that are likely to lower the cost function. Maintaining a surrogate model can be beneficial when evaluations of $c(\vec{x})$ are costly or when the cost is stochastic. For these reasons, Bayesian optimization is a natural choice for safety validation~\citep{akazaki2017causality, silvetti2017active, Deshmukh2017testing, mullins2018adaptive, abeysirigoonawardena2019generating,yang2020stress}.

Bayesian optimization iterates over two steps: 1) updating the surrogate model with new evaluations, and 2) choosing the next disturbance trajectory to evaluate. The update step computes a posterior distribution given the new evaluations (the details depend upon the employed surrogate model). Then the next disturbance trajectory is chosen using an inner optimization loop that optimizes a metric such as prediction-based exploration, error-based exploration, lower confidence bound exploration, probability of improvement, or expected improvement (see \citet{kochenderfer2019algorithms} for details). The inner optimization can be performed using simulated annealing~\citep{akazaki2017causality} or a sampling plan with good coverage of the space of disturbance trajectories~\citep{silvetti2017active}. 

One drawback to using Gaussian processes is their inability to scale to large dimensions and many sample points. To improve scalability for safety validation, \citet{Mathesen2019falsification} introduced an algorithm called stochastic optimization with adaptive restarts (SOAR) that uses a two-layered Bayesian optimization approach to trade off between global and local search. A Gaussian process model is maintained over the global search space, and a stochastic method is used to select regions for further exploration. Local Gaussian process models are used for refined searching of local regions. When a local region is no longer seeing improvement, a new region is selected. \citet{Deshmukh2017testing} addressed scaling to large dimensions using a dimensionality reduction technique based on random embeddings~\citep{wang2016bayesian}.

%%%%%%%%%%%%%%%%%%%%%%%%%%%%%%%%%%%%%%%%%%%%%%%%%%%%%%%%%%%%%%%%%%%%%%%%%%%%%%%%%%%%%%%%%%%%%%%%%%%%%%%%%%%%
\subsection{Extended Ant-Colony Optimization}
Ant colony optimization is a probabilistic technique initially used for finding optimal paths through graphs~\citep{dorigo1996ant}. Ant colony optimization was extended to continuous spaces by \citet{fainekos2003inverse} and later applied to falsification~\citep{annapureddy2010ant}. Extended ant-colony optimization works by treating each disturbance trajectory $\vec{x}$ as a path through a graph with edges between adjacent temporal points $(x_t, x_{t+1})$. At each time $t$, the space of disturbances is discretized into $N$ equally spaced cells. Ants traverse the graph by selecting their next cell $i$ at time $t$ based on the amount of pheromone present, where high pheromone implies low cost. After an ant selects their next cell, it then selects a point uniformly at random inside that cell and moves to it. At the end of each trajectory, pheromone is deposited at each cell for low-cost trajectories.

\section{Path Planning} 
\label{sec:path_planning}

This section surveys safety validation algorithms based on planning algorithms (and their required modifications). The presented algorithms are variants of the rapidly exploring random tree algorithm, multiple shooting methods, and Las Vegas tree search.

%%%%%%%%%%%%%%%%%%%%%%%%%%%%%%%%%%%%%%%%%%%%%%%%%%%%%%%%%%%%%%%%%%%%%%%%%%%%%%%%%%%%%%%%%%%%%%%%%%%%%%%%%%%%%%%%%%%%%%%%%%%%%%%%%%%%%%%%%%%%%%%%%%%%%%%%%%
\subsection{Rapidly Exploring Random Tree}
Rapidly-exploring random tree (RRT) is a path planning technique for efficiently finding failure trajectories~\citep{lavalle1998rapidly}. A tree is iteratively constructed by sampling the state space and growing in the direction of unexplored regions. RRT has been applied to the safety validation of black-box systems \citep{esposito2004adaptive,kim2005rrt,branicky2006sampling,dang2008sensitive,Nahhal2007Test,plaku2009hybrid,dreossi2015efficient,tuncali2019rapidly,koschi2019computationally}.

\begin{algorithm}
\caption{Rapidly-exploring random tree.} \label{alg:rrt}
\begin{algorithmic}[1]
    \Function{RRT}{$s_0$, $S_{\rm fail}$}
    \State $T \gets$ \textproc{InitializeTree}($s_0$)
    \Loop
        \State $s_{\rm goal} \gets$ \textproc{SampleState}() \label{line:rrt_sample_state}
        \State $s_{\rm near} \gets$ \textproc{NearestNeighbor}($T$, $s_{\rm goal}$) \label{line:rrt_nearest_neighbor}
        \State $x_{\rm new} \gets $ \textproc{GetDisturbance}($s_{\rm near}$, $s_{\rm goal}$) \label{line:rrt_optimal_input}
        \State $s_{\rm new}$ $\gets$ $f(s_{\rm near}, x_{\rm new})$ \label{line:rrt_simulate}
        \State \textproc{AddNode}($T$, $s_{\rm near} $, $s_{\rm new}$) \label{line:rrt_add_new}
    \EndLoop
    \State \Return{\textproc{CounterExamples}($T$, $S_{\rm fail}$)}
    \EndFunction
\end{algorithmic}
\end{algorithm}

% What is the basic idea of Rapidly Exploring Random Trees
The basic approach is presented in \cref{alg:rrt}. On each iteration, a random point $s_{\rm goal}$ in the state space is generated, which acts as the goal state for the next node to be added (line~\ref{line:rrt_sample_state}). The tree is searched for the node $s_{\rm near}$ that is closest to the goal state (line~\ref{line:rrt_nearest_neighbor}) based on some distance metric $d$.  This node will act as the starting point when attempting to reach the goal. A disturbance $x_{\rm new}$ is generated that drives $s_{\rm near}$ toward $s_{\rm goal}$ (line~\ref{line:rrt_optimal_input}). Since the system is a black-box, an $x_{\rm new}$ that causes $s_{\rm near} = s_{\rm goal}$ can only be approximated through random sampling or a more advanced optimization procedure. Lastly, the disturbance $x_{\rm new}$ is simulated, starting from $s_{\rm near}$, resulting in a new state $s_{\rm new}$ (line~\ref{line:rrt_simulate}) which is then added to the tree as a child of $s_{\rm near}$ (line~\ref{line:rrt_add_new}). Note that if the simulator cannot be initialized to any state, then the trace can be simulated by starting at the root and simulating the disturbances through the branch containing $s_{\rm near}$. The algorithm stops when the maximum number of iterations is reached, a suitable falsifying trajectory is found, or tree coverage reaches a specified threshold. Variants of RRT~\citep{esposito2004adaptive,kim2005rrt,branicky2006sampling,dang2008sensitive,Nahhal2007Test,dreossi2015efficient,tuncali2019rapidly,koschi2019computationally} typically differ in their approach to state space sampling, choice of distance metric for nearest neighbor selection, or by adding additional steps that reconfigure the tree for improved performance.

%% Adpative sampling
\subsubsection{Adaptive Sampling.}
When the reachable set is only a subset of the state space, a uniform sampling of goal states may be inefficient, leading to slow tree growth and low coverage values. There are several techniques for biasing goal samples toward the reachable set.

% g-RRT
An approach known as guided-RRT (g-RRT)~\citep{Nahhal2007Test} biases the selection of $s_{\rm goal}$ to regions with low coverage. In g-RRT, the state space is segmented into $n$ regions $\{ B_1, \ldots, B_n \}$ and each region is assigned a weight $w(B_i)$. Regions are sampled according to 
\begin{equation}
    P(B_{\rm goal} = B_i) = \frac{w(B_i)}{\sum_j w(B_j)} \text{,}
\end{equation}
and then $s_{\rm goal}$ is sampled uniformly from $B_{\rm goal}$. In the original version of g-RRT~\citep{Nahhal2007Test,dang2008sensitive}, the weight $w(B_i)$ is related to the increase in the lower and upper bounds of $C_{\rm star}$ when a new point is added to $B_i$. In later work~\citep{dreossi2015efficient}, the weight is $w(B_i) = \sigma(D(T, B_i))$, where $\sigma$ is the sigmoid function. The goal in both cases is to sample points that increase the coverage of the tree. 

% Approach of kim2005rrt
\citet{kim2005rrt} maintain a distribution over $s_{\rm goal}$ that has a mean biased toward low values of the cost function and a standard deviation that adaptively changes to maximize sampling in the reachable set. As the algorithm progresses, they keep track of the ratio $\beta$ of successful expansions of the tree. A successful expansion is one where 
\begin{equation}
    d(s_{\rm new}, s_{\rm goal}) > d(s_{\rm near}, s_{\rm goal}) \text{,}
\end{equation}
meaning the tree was able to grow in the direction of $s_{\rm goal}$. The ratio $\beta$ is updated after a user-specified number of iterations and is used to update the standard deviation between bounds $[\sigma_{\rm min}, \sigma_{\rm max}]$ as 
\begin{equation}
    \sigma = (1 - \beta)(\sigma_{\max} - \sigma_{\min}) + \sigma_{\rm min} \text{.}
\end{equation}
Thus, when the number of successful expansions is large, the standard deviation is reduced to continue sampling in that region, but when the tree frequently cannot grow toward $s_{\rm goal}$ the range of values is increased to better search for the reachable set. 
 
\subsubsection{Neighbor Selection.}
The choice of distance metric for finding the nearest neighbor $s_{\rm near}$ is critical for the performance of RRT. In some applications, it is acceptable to minimize the Euclidean distance between $s_{\rm near}$ and $s_{\rm goal}$~\citep{koschi2019computationally}, but this approach may be insufficient for some problems. First, Euclidean distance does not take into account whether $s_{\rm goal}$ can be reached from $s_{\rm new}$ based on the dynamics of the system. Secondly, if the reachable set is only a small subset of the state space, then points on the boundary of the reachable set will frequently be chosen as $s_{\rm near}$, limiting the coverage of the tree. Lastly, if there is a cost associated with each trajectory, it should be considered when picking which node to expand.

% Local reachability 
To encode reachability in the selection of node neighbors, \citet{kim2005rrt} use an estimate of the time to go from $s_{\rm near}$ to $s_{\rm goal}$ defined by 
\begin{equation}
    t(s_{\rm near}, s_{\rm goal}) = \begin{cases}
        d(s_{\rm near}, s_{\rm goal}) / v & {\rm if} \  v > 0 \\
        \infty & {\rm otherwise}
    \end{cases} \text{,}
\end{equation}
where the velocity is
\begin{equation}
    v = \max_{x \in X} \left. \left( -\frac{\partial d(s, s_{\rm goal})}{\partial s} f(s_{\rm near}, x) \right|_{s = s_{\rm near}} \right) \text{.}
\end{equation}
To compute the derivative $\partial d(s, s_{\rm goal}) / \partial s$ exactly requires white-box knowledge of the system, but it could be estimated with domain knowledge or from simulations~\citep{kim2005rrt}. 

% History-based selection
Another modification to neighbor selection is known as history-based selection, which penalizes the selection of nodes that fail to expand~\citep{kim2005rrt}. A node fails to expand if it is selected as $s_{\rm near}$ and creates an output $s_{\rm new}$ that is already in the tree. Failure to expand typically occurs when $s_{\rm near}$ is on the boundary of the reachable set. The number of times a node has failed to expand is stored as $n_f$ and the distance between nodes is modified as
\begin{equation}
    d_h(s_{\rm near}, s_{j})  = d(s_{\rm near}, s_{j}) + \lambda n_f
\end{equation}
for any choice of distance metric $d$ and a user specified constant $\lambda > 0$. \citet{kim2005rrt} chose $\lambda$ to balance the contribution of $d$ and $n_j$.

% Selection based on cost function. 
To incorporate a state-dependent cost function $c(s)$ into the RRT algorithm, \citet{karaman2011sampling} developed the approach known as RRT$^*$. In RRT$^*$, the neighbor selection is done by finding a set $S_{\rm near}$ of nodes in an $\epsilon$-radius of $s_{\rm goal}$ and then selecting the node in $S_{\rm near}$ that has the lowest cost, i.e.
\begin{equation}
    s_{\rm near} = \argmin\limits_{s_i \in S_{\rm near}} c(s_i)
\end{equation}
where
\begin{equation}
    S_{\rm near} = \{ s \mid d(s, s_{\rm goal}) < \epsilon \}.
\end{equation}
For the cost function,  \citet{dreossi2015efficient} use the partial robustness of the system specification while \citet{tuncali2019rapidly} use a heuristic cost for autonomous driving. 

% Modification of how the trees are maintained
\subsubsection{Other RRT Variants.}
In the basic version of RRT, a single tree is maintained and grown until termination, but under some circumstances multiple trees may be used. When searching over a space of static parameters $\theta$, a different tree can be grown for each choice of parameter in an approach called rapidly exploring forest of trees (RRFT)~\citep{esposito2004adaptive}. In RRFT, any tree that has reached a threshold coverage value or has stopped growing will be terminated, and a new tree (with a new value of $\theta$) is added to the forest. This process continues until a counterexample is found or until the parameter space has been adequately covered with fully grown trees. Another example of maintaining multiple trees is the approach of \citet{koschi2019computationally} where a fixed number of nodes $K$ are added at each iteration, including the first (so $K$ trees are maintained). Each new node is added to the tree that has the closest node. Trees that are not being grown can be removed for memory efficiency~\citep{koschi2019computationally}.

\citet{tuncali2019rapidly} combined stochastic global optimization techniques with RRT by including a transition test (similar to simulated annealing) and a similarity test for the addition of new nodes. New nodes are only added if they pass the transition test (based on their cost function) and if they are sufficiently different from all of the other nodes in the tree. These two techniques enhance exploration and coverage of the state space. 

\citet{koschi2019computationally} introduced the backwards algorithm for RRT which connects nodes in the tree backward in time without breaking the black-box assumption. The algorithm starts by sampling a state $s_0$ in the failure set. At each iteration, a state is randomly sampled as $s_{\rm goal}$, and the nearest neighbor $s_{\rm near}$ in the tree is determined. A disturbance is selected that drives $s_{\rm goal} $ toward $s_{\rm near}$ (notice this is opposite from the basic algorithm), and a new state $s_{\rm new}$ is generated by simulating $f(s_{\rm goal}, x_{\rm new})$. To maintain continuity in the tree, $s_{\rm new}$ replaces $s_{\rm near}$ and the entire branch connecting $s_{\rm near}$ to $s_0$ is re-simulated with the same disturbances. If the resulting state (the new $s_0$) is no longer in the failure set, then the branch is terminated and the algorithm repeats. This approach, although more computationally expensive, showed improved performance for finding rare failure events of an adaptive cruise control system~\citep{koschi2019computationally}. Note, however, that unlike the basic RRT algorithm, this approach requires a simulator that can be initialized based only on the observed state of the environment.

%%%%%%%%%%%%%%%%%%%%%%%%%%%%%%%%%%%%%%%%%%%%%%%%%%%%%%%%%%%%%%%%%%%%%%%%%%%%%%%%%%%%%%%%%%%%%%%%%%%%%%%%%%
\subsection{Multiple Shooting Methods}
Multiple shooting methods~\citep{bock1984multiple} solve non-linear initial value problems and have been used for robotic motion planning~\citep{diehl2006fast} and falsification~\citep{zutshi2013trajectory,zutshi2014multiple}. The idea is to sample trajectory segments using random initial states and random disturbances. A shortest path problem is solved to find a \emph{candidate trajectory} through these segments that connects initial states and failure states, where two segments are connected if the terminal state of one segment is sufficiently close to the starting state of another. The discovered path may not be feasible due to the gaps between segments, so an optimization routine is used to find disturbances that minimize those gaps and find an actual failure trajectory. In most multiple-shooting algorithms, the procedure to connect trajectory segments involves using white-box information such as dynamical equations or gradients~\citep{zutshi2013trajectory}, and therefore multiple-shooting algorithms are not in general applicable to the black-box setting.

\citet{zutshi2014multiple} introduces two ideas that make multiple shooting methods applicable for black-box falsification in large state spaces~\citep{zutshi2014multiple}:
\begin{enumerate}
    \item The state space should only be explored around the reachable set.
    \item State space discretization and refinement can be used to connect segments.
\end{enumerate}
To achieve this, the state space is implicitly discretized into disjoint cells $C$, each of size $\delta$. The cells are not explicitly stored, but they are constructed so it is efficient to find which cell contains a given state. One efficient representation is a Cartesian grid with a fixed cell size. The reachable set is estimated by running simulations forward in time with random disturbances and recording which cells are connected together. The cell connections form a graph that can be searched for candidate falsifying trajectories. Cells in these trajectories are refined until an actual counterexample is found. Since the algorithm relies on a graph search over a discretized grid, it may be useful to provide a set of initial states $S_0$ to have more than one starting cell. 
 
\begin{algorithm}
\caption{Black-box multiple shooting method.} \label{alg:multiple_shooting}
\begin{algorithmic}[1]
    \Function{MultipleShooting}{$S_0$, $S_{\rm fail}$, $\delta$, $\gamma$,  $N_{\rm max}$, $K$}
    \State $T \gets \emptyset$ \label{line:ms_trajectories_init}
    \Loop
        \State $G \gets \textproc{SampleSegments}(S_0, T, \delta, N_{\rm max}, K)$ \label{line:ms_sample_segments}
        \State $T \gets \textproc{FindCandidateTrajectories}(G, S_0, S_{\rm fail})$ \label{line:ms_find_candidate_traj}
        \If {$T = \emptyset$}
            \State \Return{$\emptyset$}
        \EndIf
        \If{$\textproc{ActualTrajectories}(T) \neq \emptyset$} \label{line:ms_actual_trajectories}
            \State \Return{$\textproc{ActualTrajectories}(T)$}
        \EndIf
        \State $\delta \gets \gamma \delta$ \label{line:ms_reduce_gridsize}
    \EndLoop
    \EndFunction
    
    \Function{SampleSegments}{$S_0$, $T$,  $\delta$, $N_{\rm max}$, $K$}
        \State $G \gets \emptyset$
        \State $Q \gets \textproc{SampleCells}(S_0, \delta)$ \label{line:ms_sample_cells}
        \While{$Q \neq \emptyset$ \textbf{and} $|G| < N_{\rm max}$}
            \State $C \gets \textproc{Pop}(Q)$ \label{line:ms_pop_cell}
            \State Sample $\{s_1, \ldots s_K \}$ uniformly from $C$ \label{line:ms_sample_state}
            \State Sample $\{x_1, \ldots, x_K \}$ from $p(x)$  \label{line:ms_sample_disturbance}
            \For{$i \in 1:K$}
                \State $s_{\rm new} = f(s_i, x_i)$ \label{line:ms_sim}
                \State $C_{\rm new} \gets \textproc{FindCell}(s_{\rm new}, \delta)$ \label{line:ms_new_cell}
                \If{$C_{\rm new} \in T$ \textbf{or} $T = \emptyset$} \label{line:ms_check_cell_in_traj}
                    \State $G \gets G \cup (C, C_{\rm new}, x_i)$ % TODO: Set notation with {} instead of ()?
                \EndIf
            \EndFor
        \EndWhile
        \Return{G}
    \EndFunction
\end{algorithmic}
\end{algorithm}
 
The approach is presented in \cref{alg:multiple_shooting}. It takes as input a set of initial states $S_0$ and a set of failure states $S_{\rm fail}$, a discrete cell size $\delta$, a refinement factor $\gamma$, a maximum number of segments per iteration $N_{\rm max}$, and the number of samples per cell $K$. A set of candidate trajectories $T$ is initialized to the empty set (line $\ref{line:ms_trajectories_init}$). On each iteration, a graph $G$ is constructed with edges $(C_i, C_j, x_{ij})$, where $x_{ij}$ is the disturbance that transformed the system from a state $s_i \in C_i$ to a state $s_j \in C_j$ (line \ref{line:ms_sample_segments}). An initially empty graph is constructed as follows:
\begin{itemize}
    \item Some starting cells are sampled by sampling states in the initial set $S_0$, finding the corresponding cell, and adding the cell to a queue $Q$  (line \ref{line:ms_sample_cells}).
    \item On each iteration, a cell $C$ is popped off the queue and $K$ states are sampled uniformly at random from within $C$ (line \ref{line:ms_sample_state}). Then, a disturbance is randomly sampled for each state (line \ref{line:ms_sample_disturbance}).
    \item Each sampled state $s_i$ is simulated forward one timestep based on the corresponding random disturbance $x_i$ (line \ref{line:ms_sim}), resulting in a new state $s_{\rm new}$ in cell $C_{\rm new}$ (line \ref{line:ms_new_cell}).
    \item To ensure the search focuses on promising regions of the state space, the edge ($C$, $C_{\rm new}$, $x_i$) is only added to the graph if $C_{\rm new}$ is in the set of candidate trajectories $T$ (line \ref{line:ms_check_cell_in_traj}). On the first iteration, when the set of candidate trajectories is empty, all edges are added.
\end{itemize}

Once the graph is constructed, it is searched for candidate trajectories that connect cells in the initial set to cells in the failure set (line \ref{line:ms_find_candidate_traj}). If no candidate trajectories are found, then the algorithm failed. Each candidate trajectory in $T$ is simulated from its starting state using the disturbance applied at each segment to get an actual trajectory (line \ref{line:ms_actual_trajectories}). The actual trajectory will not match that candidate trajectory, since the candidate trajectory is made up of disjoint segments. If the actual trajectory is a counterexample, then return it, otherwise, reduce the grid size by a factor of $\gamma$ (line \ref{line:ms_reduce_gridsize}) and repeat the procedure.

 %%%%%%%%%%%%%%%%%%%%%%%%%%%%%%%%%%%%%%%%%%%%%%%%%%%%%%%%%%%%%%%%%%%%%%%%%%%%%%%%%%%%%%%%%%%%%%%%%%%%%%%%%%%%
\subsection{Las Vegas Tree Search}
\label{subsec:lvts}
Las Vegas tree search (LVTS)~\citep{ernst2019fast} is a tree-based falsification algorithm based on two ideas:
\begin{enumerate}
    \item The disturbance trajectory $\vec{x}$ should only be discretized in time as finely as it needs to be.
    \item The system is more likely to be falsified by extreme values of the disturbance space than less extreme values. 
\end{enumerate}
To achieve the first goal, the disturbance trajectory is constructed piecewise from disturbances of different durations. Longer duration disturbances can be implemented as a repetition of a disturbance. Let the notation $x^{[\ell]}$ indicate that $x$ is applied $\ell$ times. In LVTS, the set of possible disturbances at each step is discretized and represented by the set $Y$ and the probability of selecting disturbances $x^{[\ell]}$ is $P(x^{[\ell]})$. To achieve the second goal, $P$ is defined to favor disturbances with more extreme. LVTS (\cref{alg:lvts}) takes as input $Y$, $P$, cost function $c$, and safety threshold $\epsilon$.

\begin{algorithm}
\caption{Las Vegas tree search.} \label{alg:lvts}
\begin{algorithmic}[1]
    \Function{LVTS}{$Y$, $P$, $c$, $\epsilon$}
    \State ${\rm explored}({\vec{x}})$ $\gets$ $\emptyset$ for all $\vec{x}$ \label{line:lvts_explored_init}
    \State ${\rm unexplored}({\vec{x}})$ $\gets$ $Y$ for all $\vec{x}$ \label{line:lvts_unexplored_init}
    \Loop \label{line:lvts_outer}
        \State $\vec{x} \gets \emptyset$
        \While{${\rm unexplored}({\vec{x}}) \neq \emptyset$ or ${\rm explored}({\vec{x}}) \neq \emptyset$} 
            \State Sample $x$ from $P(x)$ \label{line:lvts_sample}
            \State $\vec{x}_{\rm new} \gets [\vec{x} \ x]$ \label{line:lvts_concat}
            \If {$x \in {\rm unexplored}({\vec{x}})$}
                \State ${\rm unexplored}({\vec{x}}) \gets {\rm unexplored}(\vec{x}) \setminus \{ x \}$ \label{line:lvts_remove_unexplored}
                \State $\underline{c}$, $\overline{c}$ $\gets$ \textproc{CostBounds}($f(\vec{x}_{\rm new}$)) \label{line:lvts_cost_bounds}
                \If $\overline{c} < \epsilon$
                    \State \textbf{return} $\vec{x}_{\rm new}$ \label{line:lvts_found_ce}
                \EndIf
                \If $\underline{c} > \epsilon$
                    \State \textbf{break}  \ref{line:lvts_outer} \label{line:lvts_restart}
                \EndIf
                \State ${\rm explored}({\vec{x}}) \gets {\rm explored}({\vec{x}}) \cup \{ x \}$
            \EndIf
            \State $\vec{x} \gets \vec{x}_{\rm new}$
        \EndWhile
    \EndLoop
    \EndFunction
\end{algorithmic}
\end{algorithm}

The algorithm proceeds by growing a tree of disturbance trajectories $\vec{x}$. Each unfinished disturbance trajectory $\vec{x}$ has associated with it a set of explored disturbances (initialized to the empty set in line \ref{line:lvts_explored_init}) and a set of unexplored disturbances (initialized to the set of all possible segments $Y$ in line \ref{line:lvts_unexplored_init}). For each iteration of the algorithm, a trajectory is stochastically expanded by sampling a new disturbance from $P$ (line \ref{line:lvts_sample}) and then concatenating it to the current trajectory (line \ref{line:lvts_concat}). If the new disturbance has not been explored before, it is removed from the unexplored set (line \ref{line:lvts_remove_unexplored}) and a lower and upper bound on the cost is computed from the function {\sc CostBounds}. If the cost function is the robustness, then the bounds can be computed by the method of \citet{dreossi2015efficient}. If the upper bound is lower then the safety threshold, then a counterexample is found and returned (line \ref{line:lvts_found_ce}). If the lower bound is greater than the safety threshold, then a counterexample is not possible for this disturbance trajectory and the algorithm restarts (line \ref{line:lvts_restart}). 

% Explain the choice of $p$
Disturbance segments are grouped by an integer $\ell \in \{1, \ldots, \ell_{\rm max} \}$ that controls the duration of the segment. For duration $\ell$, the set $Y_\ell$ is given by
\begin{equation}
    Y_\ell = \{x^{[\ell]} \mid x_i = x_{i, \min} + \alpha_{i,j} (x_{i, \max} - x_{i, \min})\} \text{,}
\end{equation}
where $x_i$ is the $i$th component of the disturbance space and $x_{i, \min}$ and $x_{i, \max}$ are the minimum and maximum values of that component. The factor
\begin{equation}
\alpha_{i,j} = (2j + 1) / 2^{b_i}
\end{equation}
for any integer $j < (2^{\ell_{\rm max} - \ell}-1) / 2$ and  $b_1 + \ldots + b_n = 1$. The construction of these sets ensures that segments with larger values of $\ell$ have longer duration and more extreme disturbance values.  The full set of segments $Y$ is the union of all the $Y_\ell$ up to a maximum level of refinement $\ell_{\max}$: 
\begin{equation}
Y = \bigcup_{\ell=1}^{\ell_{\max}} Y_\ell
\end{equation}

The segment sampling distribution $P$ ensures that disturbances with larger values of $\ell$ and lower costs are chosen more often. The distribution is constructed implicitly by assigning a weight to each level of refinement
\begin{equation}
    w_\ell = \frac{|{\rm unexplored}({\vec{x}}) \cap Y_\ell| + |{\rm explored}({\vec{x}}) \cap Y_\ell|}{2^{\ell_{\rm max} - \ell}  |Y_\ell|}
\end{equation} 
and choosing $\ell$ with probability $w_\ell / \sum_{k=1}^{\ell_{\max}} w_k$. The weight is constructed to favor lower values of $\ell$ (due to the exponential factor in the denominator) until the number of unexplored edges goes to zero without finding many plausible trajectories (note that the explored set only increases if the trajectory remains plausible for falsification).  Once the refinement level is selected, one of the following four options is selected uniformly at random:
\begin{enumerate}
    \item Sample $x$ from ${\rm unexplored}({\vec{x}}) \cup Y_l$.
    \item Sample $x$ from ${\rm explored}({\vec{x}}) \cup Y_l$.
    \item Choose the $x$ from ${\rm explored}({\vec{x}}) \cup Y_l$ that minimizes the cost of $[\vec{x} \  x ]$.
    \item Choose the $x$ from ${\rm explored}(\vec{x}) \cup Y_l$ that minimizes the cost of $[\vec{x} \ x \ x^\prime]$ for all $x^\prime \in {\rm explored}([\vec{x} \ x])$.
\end{enumerate}
Option 1 ensures exploration over the unexplored set, while options 2--4 increasingly exploit knowledge of trajectories with low costs. 

\section{Reinforcement Learning}
\label{sec:reinforcement_learning}
The algorithms presented in this section are based on techniques from reinforcement learning. We present variants of Monte Carlo tree search and deep reinforcement learning. 

%%%%%%%%%%%%%%%%%%%%%%%%%%%%%%%%%%%%%%%%%%%%%%%%%%%%%%%%%%%%%%%%%%%%%%%%%%%%%%%%%%%%%%%%%%%%%%%%%%%%%%%%%%%%
\subsection{Monte Carlo Tree Search} 
\label{subsec:mcts}
Monte Carlo tree search (MCTS) is an online planning algorithm for sequential decision making that has seen success for long time-horizon problems~\citep{silver2016mastering}, making it useful for safety validation~\citep{lee2020adaptive,zhang2018two,wicker2018feature,delmas2019evaluation,julian2020validation,moss2020adaptive}. Monte Carlo tree search uses online planning to determine the best actions to take from a starting state to maximize reward. A search tree is maintained over all of the paths tried as well as an estimate of the value function at each step. Each iteration in MCTS consists of four steps:
\begin{enumerate}
    \item \emph{Selection}: Starting from the root of the search tree, disturbances are selected sequentially until reaching a leaf node, with the goal of choosing disturbances that are more likely to lead to high reward.
    \item \emph{Expansion}: Once the algorithm arrives at a leaf node, a new disturbance is chosen, and a node is added to the tree with zero visits.
    \item \emph{Rollout}: The value of the new node is estimated by running a simulation from the new node while choosing disturbances according to a rollout policy until the episode terminates or the finite planning horizon is reached.
    \item \emph{Backpropagation}: The value of the new node is used to update the value of all of its ancestors in the tree.
\end{enumerate}

% MCTS diagram (currently removed, for now)
% \begin{figure}[!h]
% \centering
% \resizebox{\columnwidth}{!}{\input{mcts-diagram.tex}}
% \caption{Monte Carlo tree search.}
%% TODO: Should I change the nodes to all actions? Right now, circles=states and squares=actions.
% \label{fig:mcts}
% \end{figure}

These four steps are repeated until a stopping criterion is met such as computational budget, wall-clock time limit, or a threshold for the reward of the best solution found so far. At that point, the best trace can be returned, or, for longer time-horizon problems, the best candidate disturbance is selected and the process restarts, possibly retaining the subtree associated with the selected disturbance. 

In problems where there is a discrete state or observation space, it makes sense to maintain separate state and disturbance nodes so if a state is repeated, the corresponding node in the tree can be reused, increasing the ability of the algorithm to reuse prior information. In most safety validation problems, however, the state space of the simulator will be continuous (or unavailable to the algorithm entirely), in which case the same state will rarely be sampled twice. To save memory, it is common to only include disturbance nodes in the search tree, which is equivalent to setting the state equal to the concatenation of all disturbances that led to it ($s_t = \vec{x}_{1:t}$).

The discrete nature of nodes in the search tree are incompatible with a continuous disturbance space $X$. \citet{delmas2019evaluation} discretize the disturbance space into a small number of discrete disturbances that are representative of the continuous space. As an alternative to discretization, when adding a new node to the tree, \citet{lee2020adaptive} sample a new disturbance $x_t$ from a known distribution over disturbances by selecting a random seed uniformly at random and using it to produce disturbances from $p(x)$.

% Note that a depth of only 3 or 5 is used in this work
\citet{zhang2018two} combined MCTS with global optimization, using MCTS for exploration of the disturbance trajectory space and global optimization for refinement of the disturbance trajectories. The disturbance space is first discretized into $L_1 \times \ldots \times L_n$ equal-sized regions. Each node in the tree represents one of the regions in the disturbance space denoted $B_t$. When a new node is added to the tree at depth $d$, its value is estimated by solving the following constrained optimization problem:
\begin{equation}
\begin{split}
    \max_{\vec{x}} \quad & \mathbb{E}\left[ \sum_t \gamma^t R(s_t, x_t) \right] \\ 
    \rm{s.t.} \quad & x_1 \in B_1, \quad \ldots, \quad  x_d \in B_d \text{.}
    \label{eq:MCTS_opt}
\end{split}
\end{equation}
This approach can be combined with progressive widening~\citep{coulom2007computing,chaslot2008progressive} when the disturbance space is large. Instead of randomly sampling a new region, optimization can be used to find the optimal region to expand.  When adding a new disturbance at depth $d+1$, solve \cref{eq:MCTS_opt} with the added constraint that $\vec{x}_{d+1}$ exists in the set of regions that have yet to be expanded. For ease of optimization, that subset may be further restricted to its convex subset~\citep{zhang2018two}. Once the MCTS budget is spent, traces with high reward can be found by solving the constrained optimization problem of the highest performing branch of the tree. 

The use of optimization for each new node of the search tree increases the computational cost of the algorithm. \citet{zhang2018two} note that balancing the computational budget between optimization iterations and tree expansions is critical to the success of this approach. They compare simulated annealing (SA)~\citep{abbas2013probabilistic}, Globalized Nelder-Mead (GNM)~\citep{luersen2004globalized}, and covariance matrix adaptation evolution strategy (CMA-ES)~\citep{hansen1996adapting} for the global optimization algorithms. It is noted that since CMA-ES has a built-in exploration strategy, it sees less improvement when combined for MCTS than SA and GNM. GNM, on the other hand, does not have an exploration strategy (beyond probabilistic restarts) and therefore sees a significant improvement with MCTS.

%%%%%%%%%%%%%%%%%%%%%%%%%%%%%%%%%%%%%%%%%%%%%%%%%%%%%%%%%%%%%%%%%%%%%%%%%%%%%%%%%%%%%%%%%%%%%%%%%%%%%%%%%%%%
\subsection{Deep Reinforcement Learning}
\label{subsec:drl}
Deep reinforcement learning (DRL) is a category of reinforcement learning that uses deep neural networks to represent the value function $V(s)$, the state-action value function $Q(s, x)$, or the policy $\pi(s)$. DRL has shown state-of-the-art results in playing Atari games~\citep{mnih2015human}, playing chess~\citep{silver2017mastering}, and robot manipulation from camera input~\citep{gu2017deep}. In recent years, different DRL techniques have been applied to falsification and most-likely failure analysis~\citep{Akazaki2018falsification, koren2018adaptive,corso2019adaptive,koren2019efficient,behzadan2019adversarial,kuutti2020training,koren2020adaptive,qin2019automatic}.

DRL algorithms are broadly split between value-function approaches, where the neural network is used to represent the value function, and policy search approaches, where the neural network represents the policy. There are advantages and disadvantages to both, and several algorithms are discussed below. The reader is referred to the original papers for implementation details.
 
 If the disturbance space $X$ is discrete, then the deep Q-network (DQN)~\citep{mnih2015human} algorithm can be used for falsification~\citep{Akazaki2018falsification,qin2019automatic}. In DQN, the optimal $Q$-function is estimated by a neural network that takes as input the state $s$ and outputs a value for each discrete disturbance. Disturbances are selected greedily as 
\begin{equation}
x = \argmax\limits_x Q(s, x) 
\end{equation}
with probability $1 - \epsilon$ and chosen at random with probability $\epsilon$ to encourage exploration. The parameters of the $Q$-network are updated to minimize the mean squared error between the current value and a target
\begin{equation}
Q_{\rm target}(s, x) = r(s, x) + \gamma \max_{x^\prime} Q(s^\prime, x^\prime) \text{.}
\end{equation}
Since the target value greedily selects the highest value disturbance, the $Q$-network can be trained on any sample of a state, disturbance, and reward (a feature known as off-policy learning). Many implementations of DQN therefore maintain a replay buffer that stores previous $(s, x, r, s^\prime)$ tuples which are repeatedly used to update the network parameters. This reuse of data makes DQN a relatively sample-efficient algorithm. The main drawback of DQN is that it is incompatible with large or continuous disturbance spaces. 

For large or continuous disturbance spaces, the policy itself is represented by a neural network (with parameters $\theta$) that takes the state as input and either outputs a disturbance directly (e.g. $x = \pi_\theta(s)$) or outputs parameters of a distribution from which a disturbance can be sampled (e.g. for a normal distribution $[\mu, \sigma^2] = \pi_\theta(s)$ and $x \sim \mathcal{N}(\mu, \sigma^2)$). The policy is optimized to produce higher rewards using the policy gradient method~\citep{sutton2000policy}. Policy gradient methods can suffer from high variance and can be unstable during optimization. To improve optimization stability, an approach known as trust region policy optimization (TRPO)~\citep{schulman2015trust} restricts the amount a policy can change at each step. TRPO has previously been used for falsification of autonomous vehicles~\citep{koren2018adaptive,koren2019efficient,corso2019adaptive}.

Another drawback of policy gradient methods is their inability to learn off-policy. Without data reuse, these methods can require a large number of simulations to converge. Newer approaches combine policy gradient methods with value function methods to create the actor-critic paradigm, which can perform well on problems with continuous disturbance spaces while also using previous simulation data to improve sample efficiency. Actor-critic methods use two neural networks, one for the policy (the actor network) and one for the value function (the critic network) and come in several varieties. Advantage actor critic (A2C) was used for falsification by \citet{kuutti2020training}. Its more scalable counterpart, asynchronous advantage actor critic (A3C), was used by \citet{Akazaki2018falsification}. \citet{behzadan2019adversarial} use another actor-critic method known as deep deterministic policy gradient (DDPG) combined with Ornstein-Uhlenbeck exploration~\citep{lillicrap2015continuous}. 

One potential drawback of using DRL for black-box falsification is the requirement for a simulator that can be observed after each disturbance is applied. In practice, high-fidelity simulators may be large, complex software projects, so it may be difficult to access the true simulator state at each timestep, whereas getting the results or only some partial data on the final state may be easier. \citet{koren2019efficient} developed a DRL technique (also used by \citet{corso2019adaptive}) that does not require access to the simulator state. The technique uses a recurrent neural network (RNN) with long-short term memory (LSTM) layers as the policy~\citep{hochreiter1997long}. The RNN maintains a hidden state akin to the state of the simulator that is used to make future decisions. The input to each layer is the previous disturbance and the initial state of the simulation, allowing the approach to generalize across initial conditions. 

The advantages of DRL and tree search methods can be combined in an approach called go-explore (GE)~\citep{ecoffet2019go}, which has been effective for hard-exploration problems with long time horizons and no reward shaping. GE has two phases, a tree search exploration phase, and a DRL robustification phase. While the original version of GE uses the state of the simulator when building the tree and training the robust policy, \citet{koren2020adaptive} modified the algorithm to use the history of disturbances instead, reducing the access requirements of the simulator.

\section{Importance Sampling Algorithms}
\label{sec:importance_sampling}
This section summaries algorithms for estimating the probability of failure based on importance sampling. Algorithms include the cross-entropy method, multilevel splitting, classification-based importance sampling, and state-dependent importance sampling.

\subsection{Cross-Entropy Method} \label{sec:cem}

The cross-entropy method iteratively learns the optimal importance sampling distribution from a family of distributions $q(\vec{x}; \theta)$ parameterized by $\theta$ (see ~\citet{de2005tutorial} for an overview and \citet{rubinstein2013cross} for a deeper examination). The optimal distribution parameters $\theta^*$ are found by minimizing the KL-divergence between a proposal distribution $q(\vec{x}; \theta)$ and the optimal distribution $q^*(\vec{x})$, i.e.
\begin{equation}
    \theta^* = \argmin\limits_\theta D_{\rm KL}\infdivx{q^*(\vec{x})}{ q(\vec{x}; \theta)} \label{eq:ce_min} \text{,}
\end{equation}
where $D_{\rm KL}$ calculates the KL-divergence. \Cref{eq:ce_min} can be cast as a stochastic optimization problem as 
\begin{equation}
    \theta^* \approx \argmax\limits_\theta \frac{1}{N} \sum_{i=1}^N \left[ \mathds{1}{\{ c(\vec{x}_i) < \epsilon \}} \frac{p(\vec{x}_i)}{q(\vec{x}_i; \varphi)}\log q(\vec{x}_i; \theta) \right] \label{eq:ce_max} \text{,}
\end{equation}
where $\varphi$ is any set of parameters and $\vec{x}_i$ are sampled from $q(\vec{x}; \varphi)$. \Cref{eq:ce_max} can be solved analytically when the family of algorithms is in the natural exponential family (i.e. normal, exponential, Poisson, gamma, binomial, and others), and the solution corresponds to the maximum likelihood estimate of the parameters~\citep{de2005tutorial}.

For an iterative solution to finding $\theta^*$, the initial parameters $\theta_0$ are chosen so that $q(\vec{x}; \theta_0)$ is close to $p(\vec{x})$. Then, for $k=0,1,\ldots$
\begin{enumerate}
    \item Set $\varphi = \theta_k$.
    \item Draw samples $\{ \vec{x}_1, \ldots, \vec{x}_N \}$ from $q(\vec{x};\varphi)$.
    \item Solve \cref{eq:ce_max} for $\theta_{k+1}$.
\end{enumerate}

One major challenge to this approach is the rarity of failure events. If all samples have $c(\vec{x}) > \epsilon$, then $\hat{P}_{\rm fail} =0$ and the algorithm may not converge to the optimal proposal distribution. One solution is to adaptively update the safety threshold $\epsilon$ at each iteration. At iteration $k$, a safety threshold $\epsilon_k$ and a rarity parameter $\rho$ is chosen so that the fraction of samples that have $c(\vec{x}) < \epsilon_k$ is $\rho$. The parameter $\rho$ is also known as the quantile level and is often set in the range $\rho = [0.01, 0.2]$~\citep{kim2016improving,okelly2018scalable}.

The cross-entropy method has been used to estimate the probability of failure for aircraft collision avoidance systems~\citep{kim2016improving} and autonomous vehicles~\citep{okelly2018scalable,zhao2016accelerated,huang2017accelerated}. Typically, probability distributions in the natural exponential family are used~\citep{kim2016improving,okelly2018scalable,zhao2016accelerated} so that cross-entropy updates can be performed analytically. \citet{huang2017accelerated} propose a method for using piecewise exponential distributions for more flexibility while retaining the ability to compute updates analytically. \citet{sankaranarayanan2012falsification}  discuss piecewise uniform distributions over the disturbance space, and techniques for factoring the space to reduce the number of parameters needed.

\subsection{Multilevel Splitting}\label{subsec:multilevel_slitting}

Multilevel splitting~\citep{kahn1951estimation} is a non-parametric approach to estimating the optimal importance sampling distribution. Rather than relying on an explicit probability distribution (as in the cross-entropy method), multilevel splitting relies on Markov chain Monte Carlo (MCMC) estimation and scales better to larger dimensions~\citep{botev2008efficient}. It has been applied to the estimation of probability of failure of autonomous driving policies with a large number of parameters~\citep{norden2019efficient}.

The idea of multilevel splitting is to define a set of threshold levels $\infty = \epsilon_0  > \epsilon_1 > \ldots > \epsilon_K = \epsilon$ and assume that the probability of failure can be computed as a Markov chain of the form
\begin{equation}
    P( c(\vec{x}) < \epsilon ) = \prod_{k=1}^K P( c(\vec{x}) < \epsilon_k \mid  c(\vec{x}) < \epsilon_{k-1} ) \text{.}
\end{equation}
Given enough levels (i.e. a large enough $K$), each conditional probability 
\begin{equation} 
P_k = P( c(\vec{x}) < \epsilon_k \mid  c(\vec{x}) < \epsilon_{k-1})
\end{equation}
is much larger than $P( c(\vec{x}) < \epsilon)$ and can therefore be computed efficiently with basic Monte Carlo methods. \Cref{alg:multilevelsplitting} outlines the algorithm. At iteration $k$, $N$ samples of $\vec{x}$ are sorted by their cost function (line \ref{line:mls_eval}) and used to estimate the conditional probability $P_k$ (line \ref{line:mls_prob_est}). The total probability is updated based on the Markov assumption (line \ref{line:mls_update}). All samples with $c(\vec{x}) > \epsilon_k$ are discarded, and the remaining samples are resampled to get back up to $N$ samples (line \ref{line:mls_resample}). Those samples perform a random walk of $M$ steps using a kernel $T(\vec{x}^\prime \mid \vec{x})$ (line \ref{line:mls_mcmc}). The process repeats until the level reaches the true safety threshold. The choice of levels $\epsilon_k$ can be done adaptively~\citep{cerou2007adaptive} by selecting a rarity parameter $\rho$ such that $\rho N$ samples are kept at each iteration (line \ref{line:mls_adaptive}).

\begin{algorithm}
\caption{Adaptive multilevel splitting.} \label{alg:multilevelsplitting}
\begin{algorithmic}[1]
    \Function{MulitlevelSplitting}{$p$, $N$, $M$, $T$, $c$, $\epsilon$}
    \State $P_{\rm fail} \gets 1$
    \State $k \gets 1$
    \State $\epsilon_0 = \infty$
    \State Sample $\{\vec{x}_1, \ldots, \vec{x}_N \}$ from $p(\vec{x})$
    
    \While{$\epsilon_k > \epsilon$}
        \State $k \gets k+1$
        \State Sort $\{\vec{x}_1, \ldots, \vec{x}_N\}$ by $c(\vec{x}_i)$ \label{line:mls_eval}
        \State $\epsilon_k \gets \max(c(\vec{x}_{\rho N}), \epsilon)$ \label{line:mls_adaptive}
        \State $P_k \gets \frac{1}{N} \sum_{i=1}^{N} \mathds{1}{ \{ c(\vec{x}_i) < \epsilon_k \} }$ \label{line:mls_prob_est}
        \State $P_{\rm fail} \gets P_{\rm fail} P_k$ \label{line:mls_update}
        \State  Resample $\{\vec{x}_1, \ldots, \vec{x}_N \}$ from $\{\vec{x}_1, \ldots, \vec{x}_{\rho N} \}$ \label{line:mls_resample}
        \State Random walk $M$ steps using $T(\vec{x}^\prime \mid \vec{x})$ for each $\{\vec{x}_1, \ldots, \vec{x}_N \}$ \label{line:mls_mcmc}
    \EndWhile
    \State \textbf{return} $P_{\rm fail}$
    \EndFunction
\end{algorithmic}
\end{algorithm}

\subsection{Classification Methods}
Supervised learning can be used to classify disturbances as safe or unsafe. That classification can then be combined with importance sampling to estimate the probability of failure of the system. Supervised learning often requires observing more than one example of a failure, so these approaches are most applicable when failures can be found relatively easily. One approach is to use a space-filling sampling plan for the disturbance~\citep{huang2018versatile} to ensure good coverage of the disturbance space, but this fails to scale to high dimensions. Another approach is to use previous versions of the system that are far less safe~\citep{uesato2019rigorous}. Known as a \emph{continuation approach}, this technique is well suited to black-boxes that have learned behavior. During the learning process, the system will fail more easily (but in related ways) to the version that is ultimately being tested. Therefore, earlier versions of the system can be used for classification of disturbances.

One way to combine supervised learning with importance sampling was explored by \citet{huang2018versatile}. They built a proposal distribution centered on the boundary between safe and unsafe disturbances. If $q(\vec{x}; \theta)$ is a proposal distribution with parameters $\theta$, then they search for the point $\vec{x}^*$ in the set of failures that maximizes the probability of the proposal distribution: 
\begin{equation}
    \vec{x}^* = \argmax\limits_{\vec{x}} \mathds{1}{\{c(\vec{x}) < \epsilon\}} q(\vec{x}; \theta) % \text{.}
\end{equation}
Then, they adjust the parameters of the proposal distribution such that the mean is located at $\vec{x}^*$. This approach has shown to construct an efficient proposal distribution when the unsafe set is convex~\citep{sadowsky1990large}. In practice, it is unlikely that the set is convex, but a linear boundary between safe and unsafe examples can be constructed using the support vector machine (SVM) algorithm~\citep{suykens1999least} if the disturbance space is lifted to a higher dimension through a mapping $\phi(\vec{x})$.

Once a boundary is found, the choice of distribution $q$ determines the feasibility of computing $\vec{x}^*$. One approach is to represent $q$ by a mixture of $K$ Gaussians with weights $\alpha$: \begin{equation}
    q_{\rm GMM}(\vec{x}; \mu_i, \sigma_i) = \sum_{i=1}^{K} \alpha_i \mathcal{N}(\vec{x}; \mu_i, \sigma_i)
\end{equation}
The optimal mean for each Gaussian model $\vec{x}^*_i$ is determined separately and the proposal distribution is reconstructed by
\begin{equation}
    q^*_{\rm GMM} (\vec{x}; \vec{x}^*_i, \sigma_i) = \sum_{i=1}^{K} \alpha_i \mathcal{N}(\vec{x}; \vec{x}^*_i, \sigma_i) \text{.}
\end{equation}
Note that a Gaussian mixture model (GMM) can be obtained in the lifted space $\tilde{q}_{\rm GMM}(\phi(\vec{x}))$ and then reduced to the true disturbance space by integrating over the extra dimensions. The approach is outlined in \cref{alg:cbis}.

\begin{algorithm}
\caption{Classification-based importance sampling.} \label{alg:cbis}
\begin{algorithmic}[1]
    \Function{ClassificationImportanceSampling}{$p$, $M$, $\phi$, $c$, $\epsilon$}
        \State Sample $\{ \vec{x}_1, \ldots, \vec{x}_M \}$ from $p(\vec{x})$ and fit a GMM $\tilde{q}_{\rm GMM}(\phi(\vec{x}); \mu_i, \sigma_i)$
        \State Sample $\{ \vec{x}_1, \ldots, \vec{x}_N \}$ and compute  $\{ \mathds{1}{\{c(\vec{x}_1) < \epsilon \}}, \ldots, \mathds{1}{\{c(\vec{x}_N) < \epsilon \}} \}$
        \State Lift the disturbances by computing $\{ \phi(\vec{x}_1), \ldots, \phi(\vec{x}_N) \}$
        \State Apply SVM on pairs $\{ (\phi(\vec{x}_1), \mathds{1}{\{c(\vec{x}_1) < \epsilon \}}), \ldots , (\phi(\vec{x}_N), \mathds{1}{\{c(\vec{x}_N) < \epsilon \}})\}$
        \State Determine the optimal points $\phi^*_i$
        \State Construct $\tilde{q}^*_{\rm GMM}(\phi(\vec{x}); \phi^*_i, \sigma_i)$
        \State Marginalize to $q^*(\vec{x})$
        \State \Return{\textproc{EstimateProbability($p, q^*$)}}
    \EndFunction
\end{algorithmic}
\end{algorithm}

An alternative approach by \citet{uesato2019rigorous} uses supervised learning to estimate the probability of failure $\hat{P}_{\rm fail}(\vec{x})$ for a disturbance $\vec{x}$, and therefore does not make the assumption that failures are deterministic given $\vec{x}$. The function $\hat{P}_{\rm fail}(\vec{x})$ can be represented using a neural network or some other model and is trained on failure examples which may come from weaker versions of the system. The optimal importance sampling distribution was proven to be
\begin{equation}
    q^*(\vec{x}) = \frac{\sqrt{\hat{P}_{\rm fail}(\vec{x})} p(\vec{x})}{\mathbb{E}_{p}\left[\sqrt{\hat{P}_{\rm fail}(\vec{x})}\right]} \text{,}
\end{equation}
which can be sampled from using rejection sampling. 
  
\subsection{State-Dependent Importance Sampling}
\label{subsec:apd}
While most importance sampling approaches focus on the entire space of disturbance trajectories, a sequential decision making framework can be used to find the optimal importance sampling policy $q(x \mid s)$ for a simulation state $s$~\citep{corso2020scalable, Chryssanthacopoulos2010}. For a Markovian system and simulator, the optimal importance sampling policy is given by 
\begin{equation}
    q(x \mid s) = \frac{p(x \mid s) P_{\rm fail}(s^\prime) }{ P_{\rm fail}(s) } \text{,}
\end{equation} % TODO: p(x) on left or right (to be consistent with the q^*(\vec{x}) equations above.
where $s^\prime$ is deterministically reached after disturbance $x$ is applied in state $s$. The probability of failure $P_{\rm fail}(s)$ can be estimated through the approximate solution of the Bellman equation
\begin{equation}
    P_{\rm fail}(s) = \begin{cases}
        1 &  \text{if} \ s \in S_{\rm fail} \\
        0 &  \text{if} \  s \notin S_{\rm fail}, \ s \in S_{\rm term} \\
        \sum_a p(x \mid s) P_{\rm fail}(s^\prime) &  \text{otherwise} \text{,}
    \end{cases} \label{eq:belman}
\end{equation}
where $S_{\rm term}$ is the set of terminal states that are not failure states. Local approximation dynamic programming and Monte Carlo policy evaluations are two successful approaches for solving \cref{eq:belman}~\citep{corso2020scalable}.

\section{Problem Decomposition Techniques}
\label{sec:decomp}

One of the most significant challenges to overcome for the safety validation of autonomous systems is that algorithms often scale poorly to large disturbance spaces and state spaces. A common approach to dealing with scalability is the use of decomposition approaches to simplify a larger problem into more tractable subproblems. For a survey of decomposition approaches in MDPs see the survey of \citet{daoui2010exact}. Decomposition approaches that rely on the factorization of the transition function \citep{guestrin2003efficient} are not applicable due to the black-box assumption, but techniques that generalize over the state and action spaces could be applied to safety validation. This section will present two approaches that have been used for decomposing the falsification problem into more manageable subproblems to accelerate finding counterexamples. 

\subsection{State Space Decomposition}
The approach presented by \citet{corso2020scalable} involves decomposing the simulator state and disturbance spaces into independent components, and finding failures for each component. Originally done in the context of multiple distinct actors in a simulated driving environment, this approach can be used with any simulation that has components that can be simulated individually. To separate the state space into $M$ different components, define a decomposition operator $D$ such that
\begin{equation}
    \{ s^{(1)}, \ldots, s^{(M)} \}  = D(s)
\end{equation}
and the disturbance space $X^{(i)}$ associated with the state $s^{(i)}$ is smaller than the full disturbance space (i.e. $|X^{(i)}| < |X|$). For each subproblem, solve for a policy that finds counterexamples
\begin{equation}
    x^{(i)} = \pi^{(i)}(s^{(i)})
\end{equation}
and then combine the policies with a fusion function $F$ such that
\begin{equation}
    x = F(\pi^{(1)}, \ldots, \pi^{(M)})(s) \text{.}
\end{equation}

An approach for solving the subproblems that is amenable to policy fusion is approximate dynamic programming (see \cref{subsec:apd}). In that case, the subproblem policy is defined by computing the probability of failure for each state component $P_{\rm fail}^{(i)}$. The fusion function can then apply simple arithmetic operations (like mean, max, or min) to arrive at a joint policy
\begin{align}
    \tilde{P}_{\rm fail}(s) &= F\left(P^{(1)}_{\rm fail}(s^{(1)}), \ldots, P^{(1)}_{\rm fail}(s^{(m)})\right) \\
    x &= \frac{p(\vec{x} \mid s) P_{\rm fail}(s^\prime) }{ P_{\rm fail}(s) },
\end{align}
where $s^\prime$ is the deterministic state reached after applying disturbance $x$ at state $s$.

If the fusion function is simple, then it may not capture the joint interactions between multiple disturbance components~\citep{corso2020scalable}. To address this problem, a global correction factor can be learned from the full simulation. The estimate probability of failure is given by
\begin{equation}
    P_{\rm fail}(s) \approx \tilde{P}_{\rm fail}(s) + \delta P_\theta(s) \text{.}
\end{equation} % TODO: define \delta
The correction factor $\delta P_\theta$ is trained using rollouts $\{s_1, \ldots, s_N \}$ from the full simulation and minimizing the difference between the estimate $ \tilde{P}_\theta(s_i)$ and the actual discounted return $G(s_i)$ where
\begin{equation}
    G(s_i) = \mathds{1}\{s_N \in S_
    {\rm fail} \} \frac{\prod_{t=i}^N p(x_t \mid s_{t-1})}{\pi(x_t \mid s_{t-1})} \text{.}
\end{equation}
This approach has been shown to increase the number of failures found in a complex driving environment with a large disturbance space~\citep{corso2020scalable}.

\subsection{Compositional Approach with Machine Learning Components}
The approach of \citet{Dreossi2019compositional} is applicable when the system contains a machine-learned component that performs classification (such as a neural network based perception system). Knowing something about the structure of the system makes this approach gray-box, but due to the prevalence of ML components, it is still widely applicable. 

The algorithm is performed with the following steps:
\begin{itemize}
    \item The ML component of interest $f(\vec{x})$ is partitioned from the rest of the system and environment. The ML component, which has a large disturbance space, is replaced with an idealized abstraction with a much smaller input space.
    \item The simulator with the idealized ML component is separated into two versions: 1) $f^+(\vec{x})$ where the ML component behaves as well as possible (i.e. classifying all inputs correctly), and 2) $f^-(\vec{x})$ where the ML component behaves poorly.
    \item For each version of the simulator, a traditional falsification  algorithm is used to partition the disturbance space into the regions that satisfy the specification $\operatorname{safe}(\vec{x})$ and regions of failures $\operatorname{fail}(\vec{x})$. The notation $\operatorname{safe}(\vec{x})^\pm$ indicates the safe disturbance set for $f^\pm(\vec{x})$. Similar notation is used for the failure set. 
    \item To find counterexamples that were caused by the ML component, find falsifying examples in the set $\operatorname{safe}(\vec{x})^+ \setminus \operatorname{safe}(\vec{x})^-$. The set $\operatorname{safe}(\vec{x})^-$ is removed because no failure is possible in this region even with the ML component functioning as poorly as possible. This part of the disturbance space is referred to as the region of interest.
    \item Finally, use a separate analyzer (possibly white-box) to identify the high-dimensional inputs that lead to failures (i.e. misclassifications) in the region of interest.
\end{itemize}
This approach has been able to find counterexamples in a neural network perception system used by an autonomous vehicle~\citep{Dreossi2019compositional}. A similar approach by \citet{julian2020validation} uses a white-box neural network analyzer combined with Monte Carlo tree search to find failure trajectories of an image-based control system.

% \section{Learning from Falsifying Examples}
% \label{sec:from_examples}
% \begin{itemize}
%     \item Given a falsifying trace, computes a box around it that is also falsifying using a sensitivity analysis~\citep{diwakaran2017analyzing}
%     \item Training a RNN to produce realistic failures from existing failure data~\citep{jenkins2018accident}
%     \item Heterogeneous categorization of time series. [ritchie's work]
% \end{itemize}

% sort these
% \begin{itemize}
%     \item \citet{huang2019evaluation} Separate input uncertainty from simulation uncertainty
%     \item A survey of stastical model checking is given by~\citet{agha2018survey}~\citep{agha2018survey}
%     \item Finding boundaries of autonomous system behavior using adaptive sampling and unsupervised learning~\citep{mullins2018adaptive}
% \end{itemize}

\section{Applications}
\label{sec:applications}
Autonomous cars and aircraft are two major application domains of black-box safety validation methods. While there are a variety of scenarios within these application domains, a common underlying principle is that of \emph{miss distance} as a reward heuristic. For most scenarios, it is possible to use some sort of distance metric to measure how close the system came to a failure during the scenario. This distance is then used to allow optimization to find counterexamples. The autonomous vehicles and aircraft application domains are covered in more detail below.

\subsection{Autonomous Driving}
Within the field of autonomous driving, there are multiple scenarios that are commonly used for testing:

\paragraph{Lane-Following.}
Lane-following is one of the most common examples in the autonomous vehicle field, and has been used to test systems ranging from full-stack systems~\citep{okelly2018scalable} to systems with ideal perception~\citep{behzadan2019adversarial} to systems that are not fully autonomous like advanced driver-assistance systems (ADAS)~\citep{koschi2019computationally}. The system under test tries to maintain a desired speed in the current lane and may~\citep{tuncali2019rapidly} or may not~\citep{kuutti2020training} have the ability to change lanes. The goal is often to find collisions, most commonly rear-end collisions. Scenario variations include both highway driving~\citep{norden2019efficient} and local road driving~\citep{tuncali2019requirements}.

\paragraph{Intersection Scenarios.}
In an intersection scenario, the system under test approaches a stoplight~\citep{tuncali2019requirements}, stop sign~\citep{abeysirigoonawardena2019generating}, crosswalk~\citep{koren2018adaptive}, or other form of intersection and must proceed through without a failure. Failures can include collisions with pedestrians~\citep{koren2020adaptive} or other vehicles~\citep{tuncali2019requirements}. Failures may also include violations of traffic laws~\citep{kress2008automatically} or other rule-sets, such as those designed to prevent at-fault collisions~\citep{hekmatnejad2020search}.

\paragraph{Lane Change Scenarios.}
Lane change scenarios can involve the test vehicle initiating a lane change or reacting to one~\citep{zhao2016accelerated,qin2019automatic}. %is this less clumsy?
%The system under test must either change lanes or an actor in the scenario is changing lanes into the system under test's lane~\citep{zhao2016accelerated}. 
Existing work has considered ADAS and other driver aid systems~\citep{huang2018versatile}. Failure modes include rear-end collisions as well as side collisions caused by turning into an occupied lane.

\paragraph{Platooning Vehicles.} \citet{hu2000towards} present a platooning scenario, where a number of cars are following a lead car while keeping a specific trailing distance. The vehicles are subjected to various stochastic disturbances arising from road conditions, wind conditions, or the presence of human operators. The goal of the system is to minimize the time that platoon vehicles spend in ``chasing'' mode, where they attempt to catch up to the vehicle ahead. The system fails when any of the vehicles gets too far from or too close to the leading vehicle.

\paragraph{} Within these applications, researchers have developed domain specific techniques to improve performance. The most common approach is to use miss distance, the physical distance between the test vehicle and other vehicles in the simulation, as a reward heuristic to allow easier optimization~\citep{koren2020adaptive}. Miss distance is also the underlying principle for robust semantics of temporal formulas when applied to autonomous vehicles~\citep{zhang2018two}. Some work has been done on identifying situations where a collision is imminent instead of the collision itself, sometimes called ``unsafe states''~\citep{koschi2019computationally} or ``boundary cases''~\citep{tuncali2019rapidly} since these regions may be easier to find. 

\subsection{Autonomous Flying and Aircraft Collision Avoidance}
Black-box validation techniques have also been applied to aircraft in multiple ways:

\paragraph{Flight Control Software.}
\citet{delmas2019evaluation} present an application where the controller must keep an aircraft in steady flight in response to disturbances such as wind or pilot inputs. Failures include reduced flight quality, autopilot disengagements, and overshoots of expert-defined thresholds. A second application is presented by \citet{ernst2019arch}, where an F-16 controller performs automatic maneuvers to avoid ground collisions. Validation algorithms search for violations of a minimum altitude from various initial conditions. \citet{julian2020validation} perform safety validation on a neural network controller with camera inputs for aircraft taxiing.

\paragraph{Collision Avoidance.}
There are several examples of applications to collision avoidance problems for aircraft. \citet{lee2020adaptive} validate the next-generation aircraft collision avoidance system (ACAS X), which makes recommendations to pilots to avoid mid-air collisions. The system may or may not coordinate its recommendations with the other aircraft.  A failure occurs when the aircraft pass too close to one another, known as a near mid-air collision (NMAC). 
\citet{esposito2004adaptive} validate a control system that guides a group of planes from an initial location to a final destination in the presence of stochastic wind disturbances.  A failure in this case is a collision between any two aircraft. 

\paragraph{Flight Path-Planning.}
Systems are given a mission, which is a series of waypoints that the system must navigate~\citep{tuncali2018experience, moss2020adaptive}, or a target location and keep-out zones along the path~\citep{Lee2019a}. Falsifying trajectories may be different mission parameters or disturbances during mission execution, such as wind and sensor noise.  Failures occur when the system enters areas defined as off-limits, collides with an obstacle, or produces a software error. \citet{yang2020stress} validate scheduling and trajectory planning for urban air mobility and package delivery systems.

\subsection{General Systems}

Black-box safety validation has also been applied to systems that are not tied specific applications such as hybrid systems, neural network controllers and planning algorithms.

\paragraph{Hybrid Systems.} Algorithms for black-box safety validation have also been applied to a number of hybrid systems outside of autonomous driving and autonomous aircraft. \citet{Hoxha2015benchmarks} present an automatic transmission system model, which is widely used as a benchmark problem in the literature.  The system under test selects a gear based on throttle and brake inputs as well as state information such as current engine load and car speed. Failure events include violations of speed thresholds and changing gears too often. \citet{jin2014powertrain} present another widely used benchmark problem from the automotive field.  The system is an abstract fuel controller for an automotive powertrain. The controller receives inputs such as fuel-flow measurements and throttle and must output a fuel command. Failures may be either steady-state, such as the violation of an air-to-fuel ratio, or transient, such as pulses that violate a settling-time constraint. 

Non-automotive systems have also been considered as well. \citet{kim2005rrt} analyze a thermostat model under various environmental conditions and test whether the heater will be active for longer than a certain proportion of time. \citet{schuler2016hybrid} present a simplified model of a wind turbine, which attempts to generate as much power as possible based on the current wind speed. Failures are violations of safety criteria, which include violations of thresholds on tower base moment and rotor speed. 

\paragraph{Controllers.}
Systems with neural network controllers have also become common case studies for validation techniques. \citet{yaghoubi2019gray} test a steam condenser controlled by a recurrent neural network (RNN). The system modulates steam flow rate based on energy balance and cooling water mass balance. A failure occurs when the pressure falls outside the acceptable range. \citet{yaghoubi2019gray} also present a generalized non-linear system controlled by a feed-forward neural network.  A failure is defined as a constraint on the reference signal value. \citet{ernst2019arch} study a neural network controlled magnet levitation system. The controller takes a reference position as input and attempts to move the magnet to track the given reference position. A failure occurs if the magnet does not stabilize to a position close enough to the reference position within some time limit. 

\paragraph{Planning Modules.}
Falsification of planning modules is common among the autonomous vehicle and aircraft applications, but there are examples in other domains as well. \citet{kim2005rrt} present a hovercraft navigation application.  The validation task tests whether the hovercraft can successfully navigate to some goal region while subjected to stochastic wind disturbances. Similarly, \citet{zhang2018two} validate a free-floating robot system that must navigate to a desired target location. \citet{fehnker2004benchmarks} present a generalization of this task, where an agent is moving in a discrete 2D environment, sometimes called a \emph{gridworld} task. In their formulation, there are states that must be reached and states that must be avoided. A violation of either constraint is considered a failure. 

% TODO: Notes about applications: emphasize importance by calling out specific "acceptances" of black-box safety validation applications.

% \subsection{Common Benchmarks}
% \begin{itemize}
%     \item The 2019 ARCH competition report on falsification~\citep{ernst2019arch}
%     \item Suggested standard benchmarks and specifications to test for automotive systems (automatic transmission and fault-tolerant field control system)~\citep{Hoxha2015benchmarks}
% \end{itemize}

% \subsection{Autonomous Driving}   
% \begin{itemize}
%     \item RSS [\todo{citation}] rules encoded as STL~\citep{hekmatnejad2019encoding}
%     \item Full stack AV simulator that implemented Simulated Annealing falsification~\citep{abbas2017safe} \todo{Check if there is a new algorithm in this paper}
% \end{itemize}

% \subsection{Autonomous Flying and Aircraft Collision Avoidance}
% \begin{itemize}
%     \item Application of S-TaLiRo (various falsification techniques) to UAV simulator~\citep{tuncali2018experience}
%     \item ACASX probability of failure estimation~\citep{kim2016improving}
%     \item UAV sense and avoid~\citep{zou2014safety}
% \end{itemize}
\section{Existing Tools}
\label{sec:tools}
Implementations of various safety validation algorithms have been made available as tools for others to use.
The existing tools range from open-source academic-based toolboxes to closed-source commercial software.
A detailed survey that includes falsification tools is provided in \citep{kapinski2016simulation}.
Each of the tools described in this section create falsifying disturbance trajectories to a system given a set of system requirements that should be satisfied.
Certain tools also perform most-likely failure analysis and failure probability estimation.
Many of the existing tools interface with the MATLAB programming language/environment to stress industry standard Simulink models. 
The tools surveyed in this section focus on black or gray-box testing of cyber-physical systems and although some of these tools include additional functionality, this section focuses on features of the safety validation components.
A brief overview of the existing safety validation tools will be discussed and their benefits and restrictions will be highlighted.

\subsection{Academic Tools}
Many of the existing safety validation tools are products of academic research and released as experimental prototypes.
Although these tools are prototypes, several have matured enough to gain wide-spread usage and acceptance~\citep{diwakaran2017analyzing, tuncali2018experience,zutshi2014multiple, dreossi2015efficient,zhang2019multi}.
Two particular falsification tools have become benchmark standards in the field: \staliro{} \citep{fainekos2019robustness} and Breach \citep{donze2010breach}.
Both are open-sourced MATLAB toolboxes for optimization-based falsification.
% Both of these toolboxes are written in MATLAB and are open-sourced.
While most of the tools are open-sourced, two other tools referenced in the falsification literature are not publicly available but still discussed.
Although their respective papers indicate how to recreate their work, none of those tools have code or executables available online.
The following collection is organized into optimization-based and reinforcement learning-based tools.

\subsection{Optimization-based Tools}
The tools in this section employ a standard optimization-based technique to search for counterexamples.
\Cref{sec:optimization} outline the approaches implemented by the following tools.

\paragraph{\staliro.} \staliro{} (Systems Temporal Logic Robustness) \citep{annapureddy2011staliro,fainekos2019robustness} is a simulation-based MATLAB toolbox for temporal logic falsification of non-linear hybrid systems.
\staliro{} parameterizes the disturbance space to reformulate the falsification problem as an optimization problem.
\staliro{} instructs the user to specify system requirements in temporal logic formulas and then constructs the optimization cost function to minimize a global robustness metric.
% Input values and ranges are provided by the user and a parameterized number of control points is used to construct the input signals.
Various optimization techniques are included in the \staliro{} toolbox, such as simulated annealing (described in \cref{sec:simulated_annealing}), genetic algorithms (described in \cref{sec:genetic_algorithms}), stochastic optimization with adaptive restarts (described in \cref{subsec:bayes_opt}), the cross-entropy method (described in \cref{sec:cem}), and uniform random sampling.
% which are described in \cref{sec:simulated_annealing} and \cref{TODO} \todo{Which reference are you looking for here?}.
\staliro{} is designed to analyze arbitrary MATLAB functions or Simulink/Stateflow models.
% , thus is limited to the MATLAB environment.
\staliro{} is open-source and available under the GNU General Public License (GPL).\footnote{\url{ https://sites.google.com/a/asu.edu/s-taliro}}
% \todo{Mention ``black-box-ness''? All the tools support this, but with the assumption that the system is a model in MATLAB.}

Specific add-ons to \staliro{} have been implemented that extend the core falsification functionalities.
These add-ons generally provide other solution methods or provide additional simulation environments that interface with \staliro.
% Two applicable add-ons are described as follows.

% Removed DP-TaLiRo based on Feinekos suggestion
% \subparagraph{\dptaliro.} \dptaliro{} (Dynamic Programming Temporal Logic Robustness) \citep{yang2013dynamic} is an algorithmic add-on to \staliro{} that can be used as the robustness computation engine.
% A dynamic programming algorithm is implemented for the robustness metric.
% \dptaliro{} is included as a standard part of \staliro{}.

% Removed Sim-ATV.
% \subparagraph{Sim-ATAV.} Sim-ATAV (Simulation-based Adversarial Testing for Autonomous Vehicles) \citep{tuncali2019requirements} is an add-on to \staliro{} that provides a simulation framework for autonomous vehicle testing. Sim-ATAV is an open-source Python framework and is available under the MIT license.\footnote{\url{https://sites.google.com/a/asu.edu/s-taliro/sim-atav}}

% Breach: Optimization-based sensitivity analysis of input parameter-space.
\paragraph{Breach.} Breach~\citep{donze2010breach,Dreossi2019compositional} is a simulation-based MATLAB toolbox 
for falsification of temporal logic specifications for hybrid dynamical systems, similar to \staliro.
Breach uses optimization-based techniques including simulated annealing (described in \cref{sec:simulated_annealing}), genetic algorithms (described in \cref{sec:genetic_algorithms}), globalized Nelder-Mead~\citep{luersen2004globalized}, and CMA-ES~\citep{hansen1996adapting}. 
% Optimization techniques: 
% simulated annealing
% globalized Nelder-Mead
% covariance matrix adaptation evolution strategy (CMA-ES)
% genetic algorithms
%% The two primary components of Breach are reachability using sensitivity analysis and property-driven parameter synthesis. The reachable set of trajectories in a system is approximated using sensitivity analysis. \todo{Is sensitivity analysis required? If so we need to say a little bit more as to why we're are including this tool} The iterative reachability process can be run until approximations of trajectories are refined enough. A local iterative refinement is also used to combine sets of trajectories that satisfy a particular system property.
The user-defined system requirements are input using temporal logic formulas.
These requirements, i.e. specifications, are used to construct a cost function to be minimized based on a robustness metric.
Breach is designed to test arbitrary MATLAB functions and Simulink models and includes a MATLAB graphical user interface (GUI) that gives the user access to the input parameter sets, temporal logic formulas, and trajectory visualizations.
% , thus is limited to the MATLAB environment. 
Breach is open-source and available under the BSD license.\footnote{\url{https://github.com/decyphir/breach}}

% Gray-box, see dreossi2015efficient Section 2.1
\paragraph{\rrtrex.} \rrtrex{} (Rapidly-exploring Random Tree Robustness-guided Explorer) \citep{dreossi2015efficient} is a MATLAB falsification tool that focuses on coverage given a computational budget.
A Simulink model and user-defined requirements written in temporal logic are taken as input.
RRT path planning algorithms are used to search the disturbance space for falsifying cases, guided by a combined state space coverage metric and a robustness satisfaction  metric. \Cref{sec:path_planning} discusses the RRT approach in detail. \rrtrex{} is not currently publicly available.

\subsection{Reinforcement Learning-Based Tools}
As a direct replacement to optimization algorithms, reinforcement learning can be used as the central idea behind searching for falsifying trajectories.
The following tools implement reinforcement learning algorithms as solvers for the falsification problem.
\Cref{subsec:mcts,subsec:drl} describe the reinforcement learning algorithms implemented in the tools.

\paragraph{\falstar.} \falstar{} \citep{zhang2018two} is a prototype Scala tool for falsification of cyber-physical systems that interfaces with MATLAB through a Java API. \falstar{} uses reinforcement learning combined with optimization techniques to generate counterexamples. Techniques include Monte Carlo tree search with stochastic optimization as described in \cref{subsec:mcts} and adaptive Las Vegas tree search (aLVTS) as described in \cref{subsec:lvts}.
\falstar{} requires a Simulink model as input and uses the above techniques to generate counterexamples to the temporal logic specifications.
\falstar{} can also interface directly with the Breach toolbox to use the available solvers implemented in Breach.
\falstar{} is open-source and available under the BSD license.\footnote{\url{https://github.com/ERATOMMSD/falstar}}

\paragraph{falsify.} falsify \citep{Akazaki2018falsification} is a prototype simulation-based falsification tool that uses deep reinforcement learning.
% Typical with reinforcement learning approaches, falsify observes the change in system output behavior by controlling inputs in simulation.
Common among the academic tools, falsify can interface directly with MATLAB functions and Simulink models.
Implementing a robustness-guided approach, falsify defines the reward as a convex function of the robustness, as described in \cref{sec:reinforcement_learning}.
This robustness-guided reward function is used by two deep reinforcement learning algorithms implemented in falsify: asynchronous advantage actor critic (A3C) and double deep-Q network (double DQN), both described in \cref{subsec:drl}. The falsify tool is not currently publicly available.

\paragraph{AST Toolbox.} The AST Toolbox (Adaptive Stress Testing Toolbox) \citep{koren2018adaptive} is a Python toolbox for safety validation, which includes falsification and most-likely failure analysis.
The AST Toolbox uses reinforcement learning to find the most likely failures of black-box systems.
Two reinforcement learning techniques used as solvers are included: Monte Carlo tree search (described in \cref{subsec:mcts}) and deep reinforcement learning (described in \cref{subsec:drl}).
The AST Toolbox is built on top of two popular reinforcement learning packages, namely, OpenAI Gym \citep{brockman2016openai} and Garage.\footnote{\url{https://github.com/rlworkgroup/garage}}
Building off of these packages gives the user access to widely used reinforcement learning benchmarking problems. 

To test a system, users must provide the definitions for three basic interfacing functions.
These interfacing functions allow the tool to interact with the black-box system or simulator.
% The AST Toolbox requires the user to implement three interfacing functions to allow the tool to interact with the black-box system or simulator.
The interface defines how to \textit{initialize} the system, how to \textit{step} the system forward (returning indications of found failures or a real-valued measure of distance to a failure), and finally a means to determine if the system is in a \textit{terminal} state.
While the implementation of the interface is restricted to Python, the user can call out to existing executables or other languages from within Python.
%---i.e. wrapping the AST interface around any software callable from Python.
As an example, Python can interface with MATLAB through the MATLAB Engine API for Python.\footnote{\url{https://www.mathworks.com/help/matlab/matlab-engine-for-python.html}}
The AST Toolbox is open-source and available under the MIT license.\footnote{\url{https://github.com/sisl/AdaptiveStressTestingToolbox}}

Two related toolboxes, called AdaptiveStressTesting.jl~\citep{lee2020adaptive} and POMDPStressTesting.jl~\citep{moss2020adaptive}, follow a similar paradigm as the AST Toolbox but are implemented in the Julia programming language \citep{bezanson2017julia}.
Julia can interface directly with many other programming languages,\footnote{\url{https://github.com/JuliaInterop}} and notably, Julia can interface with MATLAB through the MATLAB.jl package.\footnote{\url{https://github.com/JuliaInterop/MATLAB.jl}}
AdaptiveStressTesting.jl\footnote{\url{https://github.com/sisl/AdaptiveStressTesting.jl}} and POMDPStressTesting.jl\footnote{\url{https://github.com/sisl/POMDPStressTesting.jl}} are open-source and available under the Apache License Version 2.0 and the MIT license, respectively.
% \todo{Mention work on POMDPStressTesting.jl?}

% \subsubsection{Closed-Source Academic Tools}
% Other tools that have been referenced in the literature but are not publicly available are discussed here. Although their respective papers indicate how to recreate their work, none of the following tools have code or executables available online.

% Other tools referenced in the literature but omitted from this survey include those which are not publicly available, e.g. \textit{falsify} \citep{Akazaki2018falsification}. 
% Although \textit{falsify} has been shown to perform well in benchmark falsification problems \citep{ernst2019arch}, its accessibility limits it as an available research tool for cyber-physical system falsification.

% ~\\
% Other tools referenced in the literature but omitted from this survey include those which are not publicly available, e.g. \textit{falsify} \citep{}. Although \textit{falsify} has been shown to perform well in benchmark falsification problems \citep{ernst2019arch}, its accessibility limits it as an available research tool for cyber-physical system falsification.

\subsection{Commercial Tools}
% Commercial software tools for safety validation of cyber-physical systems are another available option.
Certain techniques for safety validation of cyber-physical systems have become available as commercial toolboxes.
This section briefly describes these commercially available tools relating to black-box safety validation.\footnote{The authors are not affiliated with any of the companies.} 

% \paragraph{Reactis Tester.} 
% % See (Survey) "Open-Loop Testing" section.
% The Reactis Tester \citep{reactis2013} is a simulation-based MATLAB tool from Reactive Systems for automatic falsifying input generation. Reactis Tester is the first component in the Reactis toolbox used for full Simulink model verification. 
% Reactis Tester uses a patented technique called \textit{guided simulation} which uses proprietary algorithms and heuristics to generate inputs to maximize coverage for falsification.
% Reactive Systems also has a version of Reactis for C \citep{reactis2011} that has an analogous Reactis Tester for systems developed in the C programming language.
% Reactis is commercially available through Reactive Systems, Inc.\footnote{\url{https://www.reactive-systems.com/}}
\paragraph{Reactis.} 
% See (Survey) "Open-Loop Testing" section.
Reactis \citep{reactis2013} is a simulation-based MATLAB tool from Reactive Systems for falsification of Simulink models.
Reactis has three components: Reactis Tester, Reactis Simulator, and Reactis Validator.
The Reactis Tester controls the generation of falsifying trajectories.
It uses a patented technique called \textit{guided simulation} which uses proprietary algorithms and heuristics to generate trajectories to maximize coverage for falsification.
Then, the Reactis Simulator runs the system under test given the purposed falsifying trajectory.
The last component, the Reactis Validator, uses proprietary techniques to search for violations of user-defined model specifications.
Reactive Systems also has a version of Reactis for C \citep{reactis2011} that has analogous Reactis components for systems developed in the C programming language.
Reactis is commercially available through Reactive Systems, Inc.\footnote{\url{https://www.reactive-systems.com/}}

% \paragraph{SLDV.} SLDV (Simulink Design Verifier) \citep{leitner2008simulink} is a MATLAB toolbox from MathWorks for verification of open-loop controller Simulink models. SLDV exhaustively searches the user-defined input space to generate requirements-based test cases. SLDV has support for industry standards in verification including IEC certifications used internationally for system certification and DO qualifications used by the Federal Aviation Administration (FAA). This tool is included in this discussion due to its adoption as an industry standard for Simulink model verification, although test case generation is limited by exhaustive search. SLDV is commercially available through MathWorks.\footnote{\url{https://www.mathworks.com/products/simulink-design-verifier.html}} \todo{This one also seems to be kind of white box}

\paragraph{TestWeaver.} TestWeaver \citep{junghanns2008tatar} is a simulation-based falsification tool from Synopsys. TestWeaver uses proprietary search algorithms to generate falsifying trajectories while maximizing coverage. User-defined system requirements and worst-case quality indicators are used to guide the search. Extensive knowledge of the underlying system may be required to provide useful worst-case quality indicators, which may limit the application. TestWeaver can interface with other simulation frameworks to control the disturbance trajectories via libraries in Simulink, Modelica, Python, and C. TestWeaver is commercially available through Synopsys, Inc.\footnote{\url{https://www.synopsys.com/verification/virtual-prototyping/virtual-ecu/testweaver.html}}

% \paragraph{TrustworthySearch API.} TrustworthySearch API \citep{norden2019efficient} is a risk-based framework specifically for black-box autonomous vehicle safety validation from the software start-up Trustworthy AI.
\paragraph{TrustworthySearch API.} TrustworthySearch API \citep{norden2019efficient} is a risk-based falsification and probability estimation framework from Trustworthy AI.
TrustworthySearch API is general for black-box validation of systems and has applications in autonomous vehicle safety validation.
% TrustworthySearch API was designed for black-box autonomous vehicle safety validation.
% The tool implements a probabilistic-based approach through adaptive importance sampling techniques to efficiently search over the input space for falsification, described in \cref{sec:importance_sampling}.
The tool uses a proprietary sequential importance sampling and sequential Monte Carlo approach.
A version of the probabilistic-based approach was shown using adaptive importance sampling techniques described in \cref{sec:importance_sampling}.
Importance sampling is a stand-in for traditional optimization algorithms used to search the disturbance space for falsification.
The use of a proposal distribution biased towards rare failure events ensures that these low probability events are sampled more frequently.
Adaptive multilevel splitting (AMS), described in \cref{subsec:multilevel_slitting}, is implemented to estimate this biased distribution from data.
Along with falsification, the tool can also perform failure event probability estimation, unlike other commercially available products and most academic tools.
To estimate the failure probability, the probability from the biased distribution is reweighted according to the likelihood from the original unbiased distribution.
Although specific to autonomous vehicle safety validation, we include TrustworthySearch API to highlight recent advancements of black-box safety validation tools. TrustworthySearch API is commercially available through Trustworthy AI, Inc.\footnote{\url{http://trustworthy.ai/}}

\subsection{Toolbox Competition}
An academically-driven friendly competition for systems verification, called ARCH-COMP, has been held annually since 2017 \citep{ernst2019arch}. One category within the competition is the falsification of temporal logic specifications for cyber-physical systems. Falsification researchers compare their tools against common benchmark problems and use the competition to track state-of-the-art falsification tools.

As detailed in their 2019 report \citep{ernst2019arch}, \staliro{}, Breach, \falstar, and falsify participated in the most recent competition. 
Six benchmark problems from the literature were used to evaluate each tool. 
% Evaluations were run many times for each individual requirement, as some tools rely on stochastic methods.
The tools were evaluated based on their falsification rate and statistics on number of simulations required to find a falsifying trajectory.
The outcome of the 2019 competition showed that falsify had the most success, only requiring a single simulation (after training) in certain benchmark problems.
A notable emphasis on repeatability of results was made during this recent competition.
For future competitions, we suggest a comparison metric for tool runtime to help assess relative computational timing complexities.
Continuing to mature the competition as more falsification methods arise will help drive discussion around falsification tool design decisions.
The competition's benchmarks and results are available online.\footnote{\url{https://gitlab.com/goranf/ARCH-COMP}}

\subsection{Tools Discussion}
% Common among all of the available tools for falsification is an interface to MATLAB, specifically to test Simulink models. This is evident in the survey and can be attributed to an industrial emphasis on the use of MATLAB/Simulink models for prototyping. Another common component is the use of temporal logic to specify system requirements---usually specified in signal temporal logic (STL) or metric temporal logic (MTL).
% \staliro{}, Breach, and \rrtrex{} take a standard optimization-based approach to solving the falsification problem; while \falstar{}, \rrtrex{}, and falsify take a reinforcement learning approach.
The available tools can be categorized based on which aspects of the safety validation process they implement, as described in \cref{subsec:problem_formulation}: falsification, most-likely failure analysis, and failure probability estimation.
Note that all of the available tools perform falsification as their core task.
In the context of the safety validation problem, the academic tools \staliro{}, Breach, \falstar{}, \rrtrex{}, and falsify are falsification-only tools.
As for commercial tools, Reactis and TestWeaver also fall into the falsification-only category.
TrustworthySearch API is the only commercial tool that also preforms failure probability estimation and the AST Toolbox is the only academic tool that preforms most-likely failure analysis. A full comparison is outlined in \cref{tab:tools}.

\begin{table}[!t]
  \centering
  \caption{\label{tab:mcts_params} Black-box Safety Validation Tools}\label{tab:tools}
  \resizebox{\textwidth}{!}{\begin{threeparttable}
    \begin{tabular}{@{}lcccll}
    \toprule
    & \multicolumn{3}{c}{Safety Validation Problem} & & \\
    \cmidrule{2-4}
    Tool & Falsification & \shortstack{Most-Likely\\Failure} & \shortstack{Probability\\Estimation} & Technique & Source \\ % STL/MTL Req.
    \midrule
    \staliro{} \citep{annapureddy2011staliro} \tnote{$\dagger$} \phantom{ }\tnote{*} & \checkmark & \xmark & \xmark & Optimization & Open\\
    Breach \citep{donze2010breach} \tnote{*} & \checkmark & \xmark & \xmark & Optimization & Open\\
    \rrtrex{} \citep{dreossi2015efficient} \tnote{$\dagger$} & \checkmark & \xmark & \xmark & Path Planning & Closed\\
    \falstar{} \citep{zhang2018two} \tnote{$\dagger$} \phantom{ }\tnote{*} & \checkmark & \xmark & \xmark & Optimization, RL & Open\\ % TODO: Define RL somewhere.
    falsify \citep{Akazaki2018falsification} \tnote{$\dagger$} \phantom{ }\tnote{*} & \checkmark & \xmark & \xmark & Reinforcement Learning & Closed\\
    AST Toolbox \citep{koren2018adaptive} & \checkmark & \checkmark & \xmark & Reinforcement Learning & Open\\
    Reactis \citep{reactis2013} & \checkmark & \xmark & \xmark & Proprietary & Commercial\\
    TestWeaver \citep{junghanns2008tatar} & \checkmark & \xmark & \xmark & Proprietary & Commercial\\
    TrustworthySearch API \citep{norden2019efficient} & \checkmark & \xmark & \checkmark & Importance Sampling & Commercial\\
    \bottomrule
  \end{tabular}
  \begin{tablenotes}
      \item[*] {Competed in ARCH-COMP 2019 \citep{ernst2019arch}.}
      \item[$\dagger$] {Accepts system specification in temporal logic (STL or MTL).}
  \end{tablenotes}
\end{threeparttable}}
\end{table}

As for solution techniques, the tools \staliro{} and Breach take a standard optimization-based approach, while \rrtrex{} reformulates the falsification problem to solve it using path planning algorithms. 
The AST Toolbox and falsify are based on reinforcement learning and \falstar{} combines reinforcement learning and global optimization techniques for further refinement of the disturbance trajectory search.
Common among all of the tools is an interface to MATLAB, with a particular emphasis on testing Simulink models.
This is evident in the survey and can be attributed to an industrial emphasis on the use of MATLAB/Simulink models for prototyping.
A common component specific to academic falsification-only tools is the use of temporal logic to specify system requirements---encoded in signal temporal logic (STL) or metric temporal logic (MTL).
Although expressive, requiring the strict usage of a temporal logic to encode system requirements could also limit the applicability of these tools.
With a common goal of ensuring that safety-critical systems are in fact safe, the availability of these tools allows users to provide feedback given their specific use-cases and experience.
Continued tool and technique development is further encouraged by academically-driven competitions such as ARCH-COMP \citep{ernst2019arch}.
\section{Conclusion}
With the rapid increase of safety-critical autonomous systems operating with humans, it is important that we develop robust testing procedures that can ensure the safety of these systems. Due to the high level of system complexity, we generally need black-box validation strategies to find failures of the autonomous system. This work described the problems of falsification (where a single failure example is searched for), most-likely failure analysis (where the most likely failure is searched for), and failure probability estimation (where we seek a good estimate of the likelihood of failure). 

With these goals defined, we outlined a wide array of algorithms that have been used to accomplish these tasks. Global optimization, path planning, and reinforcement learning algorithms have been used to find falsifying examples, while importance sampling methods have been used to estimate the probability of failure even when it is close to zero. To address the problem of scalability, we described approaches for decomposing the safety validation problem into more manageable components. We gave a brief overview of the main applications for black-box safety validation including autonomous driving and flight. Finally, we provided an overview of the existing tools that can be used to tackle these validation tasks.
%%%%%%%%%%%%%%%%%%%%%%%%%%%%%%%%%%%%%%%%%%%%%%%%%%%%%%%%%%%%%%%%%%%%%%%%%%%%%%%%%%%%%%%%%%%%%%%%%%

\vskip 0.2in
\bibparsep=1ex
\bibhang=4ex
\printbibliography % biblatex

% \bibliography{references} % natbib
% \bibliographystyle{theapa}

\end{document}